%% file: main.tex
\ifdefined\pdfminorversion\pdfminorversion=7\fi
\PassOptionsToPackage{dvipsnames,table}{xcolor}
\documentclass{fairmeta}

\DeclareFontShape{T1}{optimistic}{b}{n}
  {<-> s * [0.88] assets/optimistic}{}

\makeatletter
\input{t1cmvtt.fd}
\makeatother
\DeclareFontShape{T1}{cmvtt}{b}{n}
  {<-> ssub * cmvtt/m/n}{}

\usepackage{amsmath,amssymb,amsthm}
\usepackage{float}
\usepackage{tabularx,ragged2e}
\usepackage{xspace}
\usepackage{xurl}
\newcolumntype{L}[1]{>{\hsize=#1\hsize\linewidth=\hsize\RaggedRight\arraybackslash}X}
\newcolumntype{C}[1]{>{\hsize=#1\hsize\linewidth=\hsize\Centering\arraybackslash}X}
\titleformat*{\paragraph}{\upshape\bfseries}

\newcommand{\low}{\textcolor{blue}{\texttt{Low}}\xspace}
\newcommand{\medium}{\textcolor{Dandelion!65!black}{\texttt{Medium}}\xspace}
\newcommand{\high}{\textcolor{Mahogany}{\texttt{High}}\xspace}
\definecolor{bestrow}{RGB}{232,242,252}
\newtheorem{theorem}{Theorem}

\newtcblisting{promptbox}[1]{
  enhanced,
  breakable,
  listing only,
  title={#1},
  colback=black!3,
  colframe=black!70,
  colbacktitle=black!70,
  coltitle=white,
  fonttitle=\small\ttfamily,
  boxrule=0.8pt,
  arc=2mm,
  left=3mm,
  right=3mm,
  top=2mm,
  bottom=2mm,
  listing options={
    basicstyle=\small\ttfamily,
    breaklines=true,
    columns=fullflexible,
    keepspaces=true,
    showstringspaces=false
  }
}
\hypersetup{pdftitle={PRIOR: Pre-Call Reasoning in SFT and Its Influence on Tool-Use Retention under RL},pdfauthor={Kaiwen Hu, Hyunin Lee, Joseph Suh, Serina Chang, Shuai Shao, Yong Zhang, Namyong Park, Xiaoqiang Luo, Luke Simon, Somayeh Sojoudi}}
\title{PRIOR: Pre-Call Reasoning in SFT and\\Its Influence on Tool-Use Retention under RL}
\author[1,*]{Kaiwen Hu}
\author[1]{Hyunin Lee}
\author[1]{Joseph Suh}
\author[1]{Serina Chang}
\author[2]{Shuai Shao}
\author[2]{Yong Zhang}
\author[2]{Namyong Park}
\author[2]{Xiaoqiang Luo}
\author[2]{Luke Simon}
\author[1]{Somayeh Sojoudi}
\affiliation[1]{University of California, Berkeley}
\affiliation[2]{Meta}
\metadata[Code]{\url{https://github.com/kaotty/PRIOR}}
\correspondence{\textsuperscript{*}Kaiwen Hu (\href{mailto:kaiwen_hu@berkeley.edu}{\texttt{kaiwen\_hu@berkeley.edu}})}

\abstract{
Tool use learned during supervised fine-tuning (SFT) is not always retained during reinforcement learning (RL). We study how properties of SFT data may affect this retention by constructing \low{}, \medium{}, and \high{} SFT datasets with
approximately matched token budgets but different tool-call depths. Under the same RL recipe, the three initializations reach similar RL training accuracy yet exhibit markedly different tool-use dynamics: tool initiation increases in \low{}, declines in \medium{}, and collapses in \high{}.
We observe that these datasets also differ substantially in the amount of reasoning preceding each tool call. In particular, \high{} trajectories contain considerably less pre-call reasoning, motivating the hypothesis that pre-call reasoning may affect how well tool use is calibrated and, consequently, whether the corresponding behaviors are retained during RL. An idealized theoretical analysis provides a selective-retention view of this process: advantages in overall and tool-conditioned accuracy can be preserved under outcome-only optimization, while tool initiation increases only when tool-using trajectories are more successful than the tool-free alternative.
Guided by this analysis, we rank the original \high{} trajectories by average pre-call reasoning length and retain the highest-ranked examples while approximately matching cumulative SFT token exposure. This simple filtering strategy improves post-RL accuracy and tool-conditioned final-answer success without consistently increasing tool initiation across multiple benchmarks. These results suggest that pre-call reasoning is a useful dimension of SFT supervision for understanding and improving \mbox{tool-use retention under downstream RL.}
}

\begin{document}
\maketitle

\begin{figure}[t]
    \centering
    \begin{subfigure}[t]{0.32\linewidth}
        \centering
        \includegraphics[width=\linewidth]{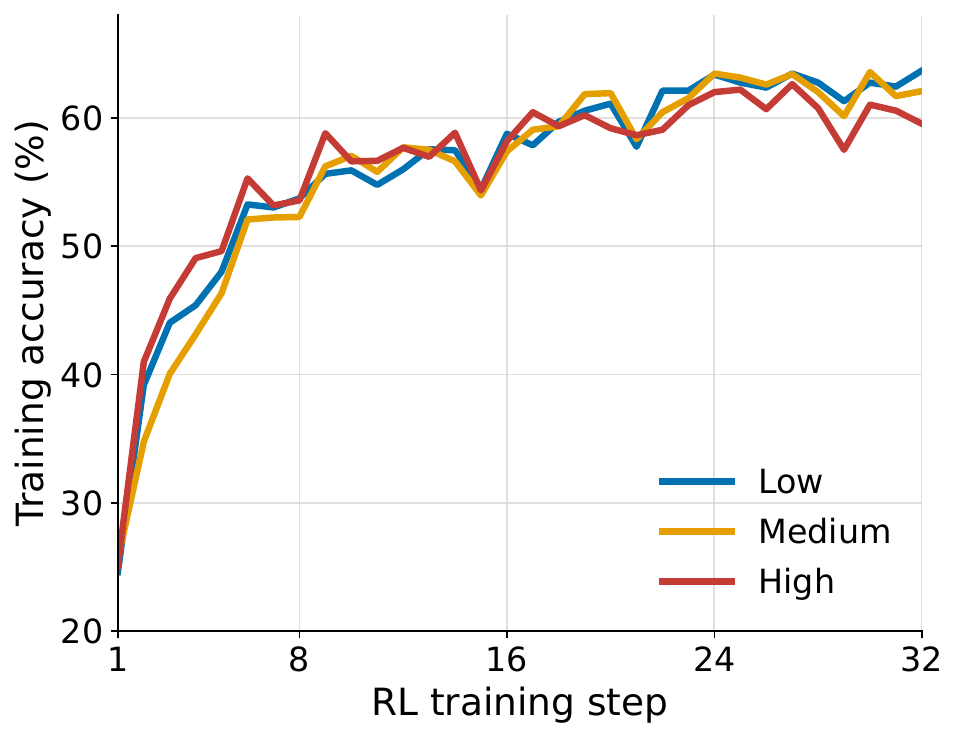}
        \caption{RL training accuracy}
        \label{fig:rl_training_acc}
    \end{subfigure}
    \hfill
    \begin{subfigure}[t]{0.32\linewidth}
        \centering
        \includegraphics[width=\linewidth]{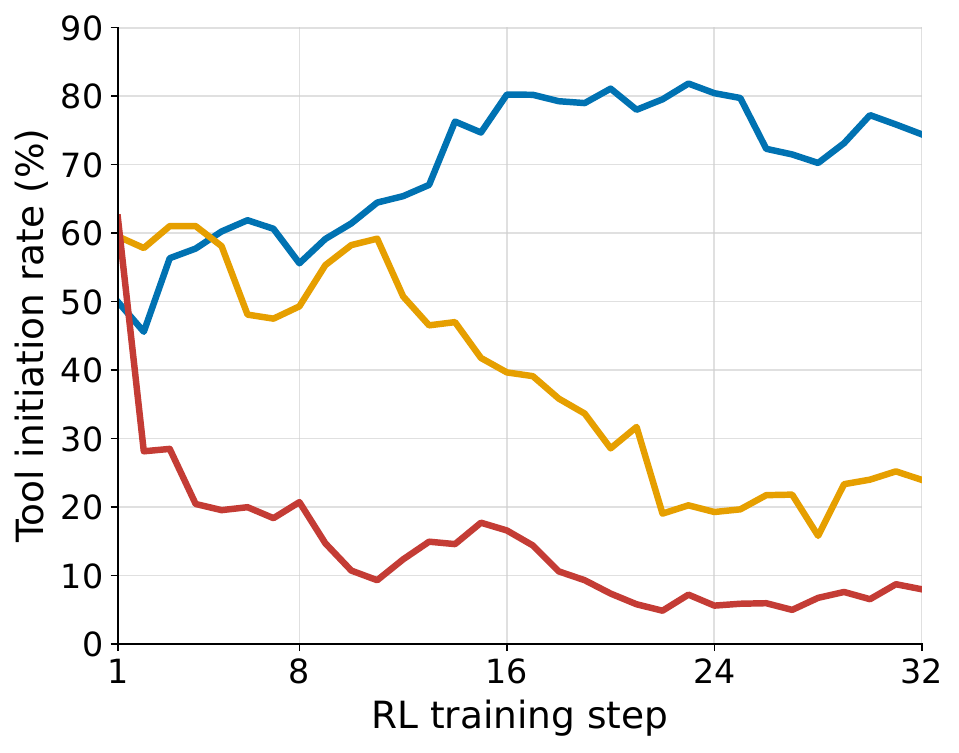}
        \caption{Tool initiation rate}
        \label{fig:rl_init_rate}
    \end{subfigure}
    \hfill
    \begin{subfigure}[t]{0.32\linewidth}
        \centering
        \includegraphics[width=\linewidth]{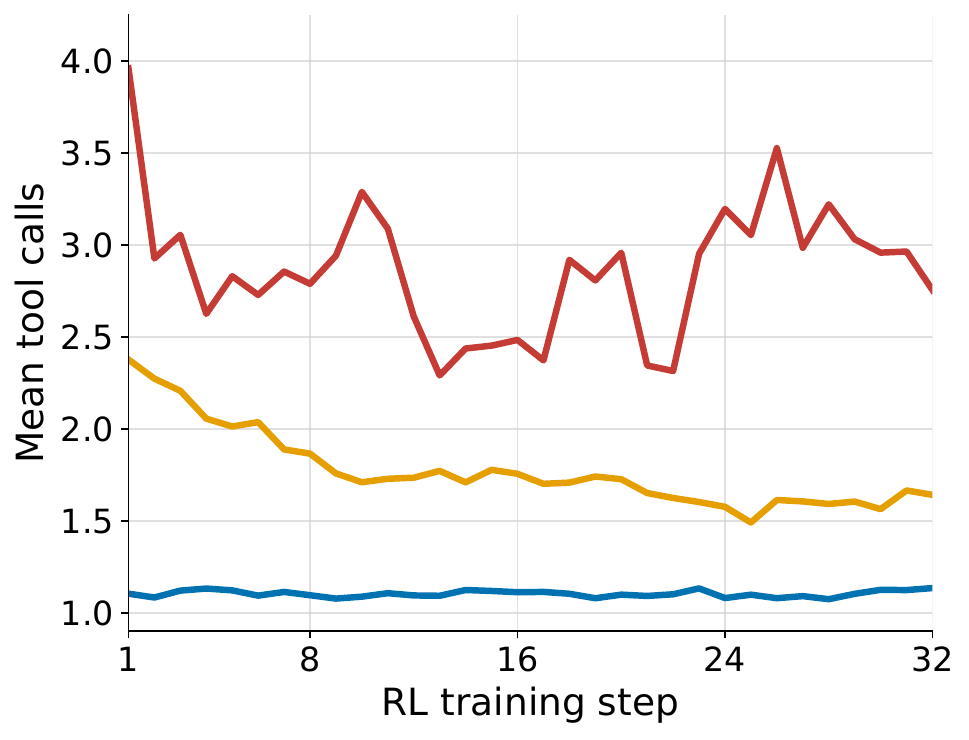}
        \caption{Conditional tool-call depth}
        \label{fig:rl_mean_tool_call}
    \end{subfigure}
    \caption{
    RL training dynamics of the \low, \medium, and \high SFT initializations. (a) RL training accuracy. (b) Fraction of rollouts that use at least one tool call within each batch. (c) Mean number of tool calls among tool-using rollouts. Labels indicate the tool-call depth of the SFT data. Despite similar training accuracy, the three initializations exhibit markedly different tool-use behavior. Under the same RL recipe, \low maintains frequent tool use, whereas \high sharply reduces tool initiation while retaining deeper interactions among the remaining tool-use trajectories.
    }
    \label{fig:rl_curve}
\end{figure}

\section{Introduction}

Recent advances in Multimodal Large Language Models (MLLMs) have demonstrated outstanding capabilities in areas such as math and coding~\citep{li2024llava,deitke2025molmo,bai2025qwen25vl,chen2024expanding}. A growing line of work further augments text-based reasoning with external tools, allowing models to zoom into image regions, execute code, retrieve information, and incorporate the resulting observations into subsequent reasoning~\citep{zheng2026deepeyes,jin2025search}. Such Tool-Integrated Reasoning (TIR) typically requires a model to decide not only \emph{how} to use a tool, but also \emph{when} it is useful and how its output should affect the remaining solution process~\citep{yao2022react,schick2023toolformer}. Recent systems commonly learn these behaviors through a two-stage pipeline: supervised fine-tuning (SFT) establishes tool-use capabilities and interaction formats, followed by reinforcement learning (RL)~\citep{hong2026deepeyesv2,song2026codedance}. While substantial effort has focused on RL algorithms~\citep{qian2026toolrl,hou2026codev,zheng2026deepeyes}, considerably less is understood about how SFT data shapes the RL training dynamics of tool-use behavior.

We investigate this question by varying the \emph{tool-call depth} of SFT trajectories, defined as the number of calls in a tool-using trajectory. Starting from the same base model with Zoom and Code tools, we construct \low, \medium, and \high SFT datasets with a matched number of trajectories and total token budgets, and then apply an identical RL training recipe. Our early experiments reveal this is a non-trivial question (Figure~\ref{fig:rl_curve}): despite similar RL training accuracy, their tool-use behaviors diverge sharply. We find that the tool initiation rate gradually increases in \low, moderately decreases in \medium, and experiences a severe collapse in \high. Among trajectories that still use tools, the depth ordering remains \high $>$ \medium $>$ \low. Thus, demonstrating more tool calls during SFT does not imply that those behaviors will be retained more frequently under outcome-only RL.

To understand this phenomenon, we examine the textual reasoning that precedes each tool call, which we term \emph{pre-call reasoning}.  \high trajectories contain substantially fewer average pre-call reasoning tokens than \low. During RL, we further observe that the tool-using trajectories that remain in \medium and \high tend to contain more pre-call reasoning after RL training than at initialization. These observations motivate the hypothesis that \emph{pre-call reasoning is a key signal for tool-use calibration}. Here, tool-use calibration refers to whether the model invokes tools when they are actually useful, rather than simply learning to call them frequently.

Our theoretical analysis formalizes this intuition through a selective-retention view of RL. In an idealized probability-space analysis, advantages in overall accuracy and tool-conditioned accuracy present in the SFT initialization are preserved under the same outcome-only optimization. Tool initiation follows different dynamics: it increases only when tool-using trajectories are more successful than the tool-free alternative, and decreases otherwise. This provides a theoretical view of selective retention: RL need not preserve frequent tool use, but instead favors tool behaviors that are better aligned with task success. Thus, tool-call frequency or depth alone is insufficient to characterize useful tool use.

Guided by this analysis, we rank the original \high trajectories by their average pre-call reasoning length and retain the highest-ranked ones. We construct two subsets containing approximately 50\% and 25\% of the original assistant-supervised token mass per epoch. Under a controlled number of cumulative supervision tokens and the same downstream RL training recipe, the 25\% variant outperforms the original \high on all five benchmarks, while the 50\% variant improves on four out of five. Moreover, the filtered initializations achieve substantially higher final-answer success among their self-selected tool-using trajectories on several benchmarks, without a consistent increase in tool initiation, suggesting that effective SFT should prioritize supervision supporting useful tool decisions rather than tool-call frequency.

Our contributions are threefold:
\begin{itemize}
    \item We show that the tool-call depth of SFT data leads to sharply different tool-use dynamics under the same outcome-only RL recipe: more frequent tool use during SFT does not necessarily translate into more persistent or better-calibrated tool use after RL.

    \item We identify pre-call reasoning as a key factor shaping \emph{tool-use calibration} and develop a selective-retention analysis showing how differences in overall and tool-conditioned success are preserved under outcome-only optimization, while tool initiation depends on whether tool use is more successful than the tool-free alternative.

    \item We show that filtering the original \high SFT data by average pre-call reasoning length, under approximately matched cumulative supervision budgets, improves downstream accuracy and tool-enabled performance without consistently increasing tool initiation, outperforming a matched random-selection control.
\end{itemize}

\section{Related Work}

\paragraph{\upshape\bfseries Tool-Integrated Multimodal Reasoning.} TIR couples language-based deliberation with external actions, allowing models to gather evidence, perform computation, and update their plans during inference. ReAct establishes the general interleaving of reasoning and acting, while Toolformer studies self-supervised learning for API selection, invocation, and result use~\citep{yao2022react,schick2023toolformer}. Recent multimodal systems extend this paradigm to visual search and executable computation: DeepEyes and VTool-R1 use RL to improve visual tool use~\citep{zheng2026deepeyes,wu2026vtool}; DeepEyesV2 and Thyme combine visual operations with code-based computation in multimodal post-training pipelines~\citep{hong2026deepeyesv2,zhang2025thyme}; and CodeDance combines task-aware cold-start SFT with difficulty-adaptive RL for executable visual reasoning~\citep{song2026codedance}. ICRL instead places demonstrations in RL prompts and gradually removes them during training~\citep{ye2026context}. These studies primarily focus on enabling multimodal tool use or improving its final performance. We study a complementary question: how the structure of SFT tool-use trajectories determines which tool behaviors remain viable under outcome-only RL training.

\paragraph{\upshape\bfseries SFT Initialization for Reinforcement Learning.} Prior work shows that SFT initialization can strongly affect subsequent RL. DeepSeek-R1 and Chu et al.~\citep{guo2025deepseek,chu2025sft} compare the roles of SFT and RL, while PEAR shows that a stronger SFT checkpoint can perform worse after identical RL~\citep{zhang2026good}. This observation closely motivates our study: SFT quality should be evaluated by the behaviors it prepares for RL, rather than by SFT performance alone. Tool-integrated studies explore cold-start data, diverse reasoning primitives, and process supervision~\citep{yao2026tailored,feng2026retool,mao2026toolvision,he2026deeptool}. ASTER reports that deep tool trajectories can provide beneficial initializations~\citep{zhang2026aster}, which appears different from our observation of tool-initiation collapse for high-depth SFT. However, ASTER emphasizes interaction density and tool-use budgets, whereas we approximately match the total SFT token budget and study how it is allocated before individual calls. Our results therefore complement this line of work by showing that the value of deep trajectories depends not only on interaction depth, but also on per-call reasoning supervision.

\paragraph{\upshape\bfseries Tool Utility and Reinforcement Learning.} Tool frequency alone does not determine tool utility. MED separates intrinsic improvement, tool-induced correction, and tool-induced harm~\citep{ma2026med}, while other work studies capability expansion, evidence dependence, and the relationship between invocation frequency and reasoning quality~\citep{guo2026multimodal,wang2026illusion,kim2026diversity}. In particular, the finding that frequent tool invocation need not improve reasoning is closely aligned with our selective-retention view~\citep{kim2026diversity}. Other approaches improve tool learning by modifying rewards, credit assignment, or rollout allocation~\citep{qian2026toolrl,hou2026codev,li2026look,xie2026tips,feng2026taco,dong2026agentic}. We instead keep outcome-only RL fixed and study how SFT data determines the reliability of the tool decisions retained by RL, separating tool frequency from the success of self-selected tool-using trajectories.

\section{How SFT Shapes Tool Use}
\label{sec:obs}

\paragraph{Notations.}
Let $\mathcal{M}$ denote an MLLM and $x$ an input prompt. Given $x$, the model interacts with the available tools and produces a trajectory $\tau \sim \mathcal{M}(\cdot \mid x)$. If $\tau$ contains $T$ tool calls, we write $\tau=(s_1,\ldots,s_T,r)$, where each \emph{tool-interaction step} $s_t$ consists of the reasoning preceding a tool call, the tool call itself, and the corresponding tool output: $s_t = (\text{reasoning}_t,\text{tool-call}_t,\text{tool-output}_t)$; $r$ is the reasoning after the last tool call that reaches the final answer. We define the \emph{tool-call depth} of $\tau$ as $d(\tau)=T$. We define $\text{reasoning}_t$ as \emph{pre-call reasoning} of $s_t$. A tool-free trajectory therefore has $d(\tau)=0$. An SFT dataset is denoted by $\mathcal{D}={(x_i,\tau_i)}_{i=1}^{N}$, where each prompt (or \emph{question}) $x_i$ is paired with its corresponding trajectory $\tau_i$. Throughout the paper, we use the terms \emph{question} and \emph{prompt} interchangeably, as well as \emph{trajectory} and \emph{rollout}, depending on the context.

We study how the tool-use patterns demonstrated during SFT affect subsequent behavior under a fixed outcome-only RL training. We vary the tool-call depth of SFT while holding the downstream training recipe fixed, and analyze both the resulting tool-use dynamics and the reasoning preceding individual tool decisions.

\subsection{Experimental Setup}
\label{sec:setup}

\textbf{Model and Tools.} We use Qwen2.5-VL-7B-Instruct~\citep{bai2025qwen25vl} as the base MLLM with two tools: \textbf{Zoom}, which crops and enlarges an image region, and \textbf{Code}, which executes model-generated Python. Models may invoke either tool multiple times. We ran all model training and evaluation experiments on two NVIDIA H200 GPUs per run.

\textbf{SFT Data Construction.}
We construct a shared source pool of tool-using Zoom and Code trajectories by standardizing prompts and tool formats and removing malformed or duplicate trajectories. Zoom trajectories are drawn from CoF~\citep{zhang2025chain}, Mini-o3~\citep{lai2026mini}, and PixelReasoner~\citep{su2026pixel}, while Code trajectories are drawn from DeepEyesV2~\citep{hong2026deepeyesv2} and Thyme~\citep{zhang2025thyme}. From this pool, we construct three SFT datasets of 2,000 trajectories each, grouped by tool-call depth: \low ($d(\tau)=1$), \medium ($2 \le d(\tau) \le 3$), and \high ($d(\tau)\ge 4$). Each dataset contains 750 Zoom trajectories, 750 Code trajectories, and the same 500 tool-free trajectories, with approximately matched trajectory lengths.

\textbf{RL Data Construction.} All SFT initializations are trained with the same fixed set of 16,384 questions with RL drawn from DeepEyes-47K~\citep{zheng2026deepeyes}, MMK12~\citep{meng2025mm}, ChartQA~\citep{masry2022chartqa}, and Geometry3K~\citep{lu2021inter}. We remove explicit tool-use hints and evaluation overlap, and use the same questions and ordering across all RL runs.

\textbf{Evaluation.} We evaluate all checkpoints on a fixed 6,431-question suite consisting of V*Bench~\citep{wu2023v} (191), ChartQA~\citep{masry2022chartqa} (2,500), MathVista testmini~\citep{lu2024mathvista} (1,000), WeMath~\citep{qiao2025we} (1,740), and CharXiv~\citep{wang2024charxiv} (1,000). We consider two evaluation settings: \texttt{tool-on} evaluation provides both Zoom and Code with up to eight calls and an 8,192-token response budget, while \texttt{tool-off} evaluation removes access to the tools while retaining the same evaluation protocol. The prompts are specified in Appendix~\ref{app:eval_prompts}.

\textbf{Training Configuration.} Each SFT model is trained for one epoch (50 optimizer updates) with assistant-only supervision. Starting from each SFT checkpoint, we run 32 updates of outcome-only GRPO~\citep{shao2024deepseekmath}, sampling five rollouts for each of 512 questions per update. RL rewards only final-answer correctness and uses the same eight-call and 8,192-token interaction budget as evaluation. All optimization settings, RL data, sampling, and decoding configurations are shared across \low, \medium, and \high. More details are provided in Appendix~\ref{app:exp_config}

\subsection{Divergent Tool-Use Dynamics under identical RL training}
\label{sec:phenomenon}

Figure~\ref{fig:rl_curve} compares the RL training dynamics of policy initialized from SFT on the \low, \medium, and \high dataset under the same RL recipe. Despite different initializations regarding the tool-call depth, all three runs finish at approximately the same training accuracy. Their tool calling behaviors, however, diverge substantially. We have three specific observations.

\paragraph{1. Tool initiation diverges across SFT initializations.} In Figure~\ref{fig:rl_init_rate}, \low increases its tool initiation rate to approximately 75\% by the end of training. In contrast, \medium exhibits an overall decline to approximately 20\%, while \high drops rapidly in the early steps and remains below 10\% near the end. Thus, greater tool-call depth in the SFT trajectories does not necessarily translate into more persistent tool initiation under the same RL recipe.

\paragraph{2. The ordering of conditional tool-call depth persists.} Figure \ref{fig:rl_mean_tool_call} describes the mean number of tool-calls among tool-using rollouts, excluding tool-free rollouts. \low consistently remains close to 1.1 tool calls, while \medium gradually decreases its tool calls from approximately 2.4 to 1.6. After a slight decline from roughly four calls, \high continues to fluctuate in this panel between 2.3 and 3.5. Thus, in contrast to the tool initiation rate, the results show that the ordering of conditional tool-call depth persists during the RL training.

\begin{table*}[t]
  \centering
  \caption{
  Evaluation results for the SFT and RL checkpoints of \low, \medium, and \high.
  Accuracy is pooled over five evaluation benchmarks.
  Gain denotes the difference between \texttt{tool-on} and \texttt{tool-off} accuracy. The three initializations exhibit \underline{similar final \texttt{tool-on} accuracy} but critically different tool-use behavior (Figure \ref{fig:rl_init_rate}).
  }
  \label{tab:eval}
  \small
  \setlength{\tabcolsep}{6pt}

  \resizebox{\textwidth}{!}{%
  \begin{tabular}{lccc|c|c}
  \toprule
  \textbf{Checkpoint}
  & \textbf{\texttt{Tool-on} Acc}
  & \textbf{\texttt{Tool-off} Acc}
  & \textbf{Gain}
  & \textbf{Tool initiation rate}
  & \textbf{Conditional tool-call depth} \\
  \midrule
  \low{} SFT    & 59.34 & 59.65 & -0.31  & 67.72 & 1.21 \\
  \low{} RL     & \underline{\textbf{67.67}} & 67.66 & +0.02  & 75.04 & 1.32 \\
  \midrule
  \medium{} SFT & 47.43 & 62.14 & -14.71 & 58.50 & 3.98 \\
  \medium{} RL  & \underline{\textbf{67.66}} & 70.56 & -2.90  & 20.82 & 3.03 \\
  \midrule
  \high{} SFT   & 43.45 & 63.52 & -20.07 & 50.60 & 5.59 \\
  \high{} RL    & \underline{\textbf{66.24}} & 68.90 & -2.66  & 5.38  & 6.32 \\
  \bottomrule
  \end{tabular}%
  }
\end{table*}

\paragraph{3. Similar accuracy can mask differences in tool-use calibration.}
Table~\ref{tab:eval} shows that RL brings the three initializations to similar final \texttt{tool-on} accuracy, despite their sharply different tool-use behaviors. \low continues to invoke tools frequently, whereas \medium and especially \high increasingly avoid them. These results show that final task accuracy alone does not reveal how tool use is being retained.
Comparing \texttt{tool-on} and \texttt{tool-off} performance shows that after RL, \texttt{tool-on} is approximately on par with \texttt{tool-off} for \low, but remains worse for \medium and \high, which indicates that the tool use in \medium and \high is not well-calibrated.
Taken together, the results suggest that outcome-only RL does not simply preserve the tool-use patterns learned during SFT: tool behaviors that are poorly calibrated to task success are less likely to be retained. This motivates our central question: what property of the SFT trajectories determines the calibration of tool use, and consequently whether that
behavior survives downstream RL?

\subsection{A Hypothesis: Pre-call Reasoning Matters}
\label{sec:hypothesis}

To investigate the divergent tool-use dynamics in Section~\ref{sec:phenomenon}, we examine how much reasoning the model performs \emph{before} each tool call, which we refer to as \emph{pre-call reasoning} (see its definition in Notations from Section~\ref{sec:obs}).

\begin{figure}[h]
    \centering
    \begin{subfigure}[t]{0.35\linewidth}
        \centering
        \includegraphics[width=\linewidth]{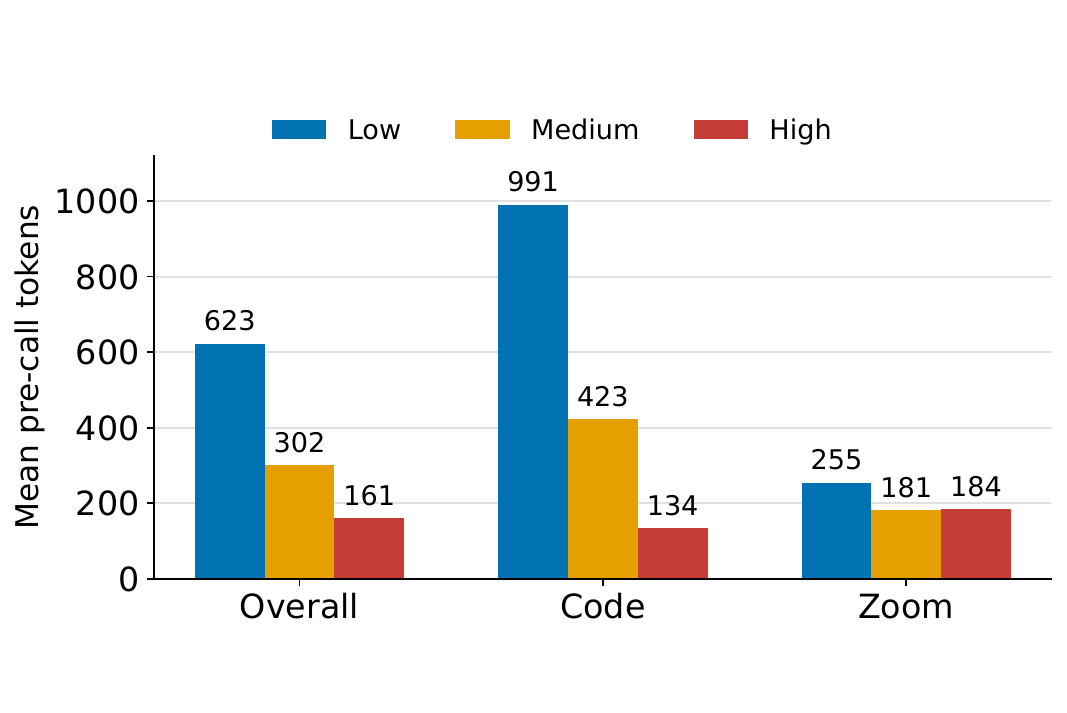}
        \caption{Mean pre-call tokens in SFT}
        \label{fig:sft_pre_call_tokens}
    \end{subfigure}
    \hfill
    \begin{subfigure}[t]{0.6\linewidth}
        \centering
        \includegraphics[width=\linewidth]{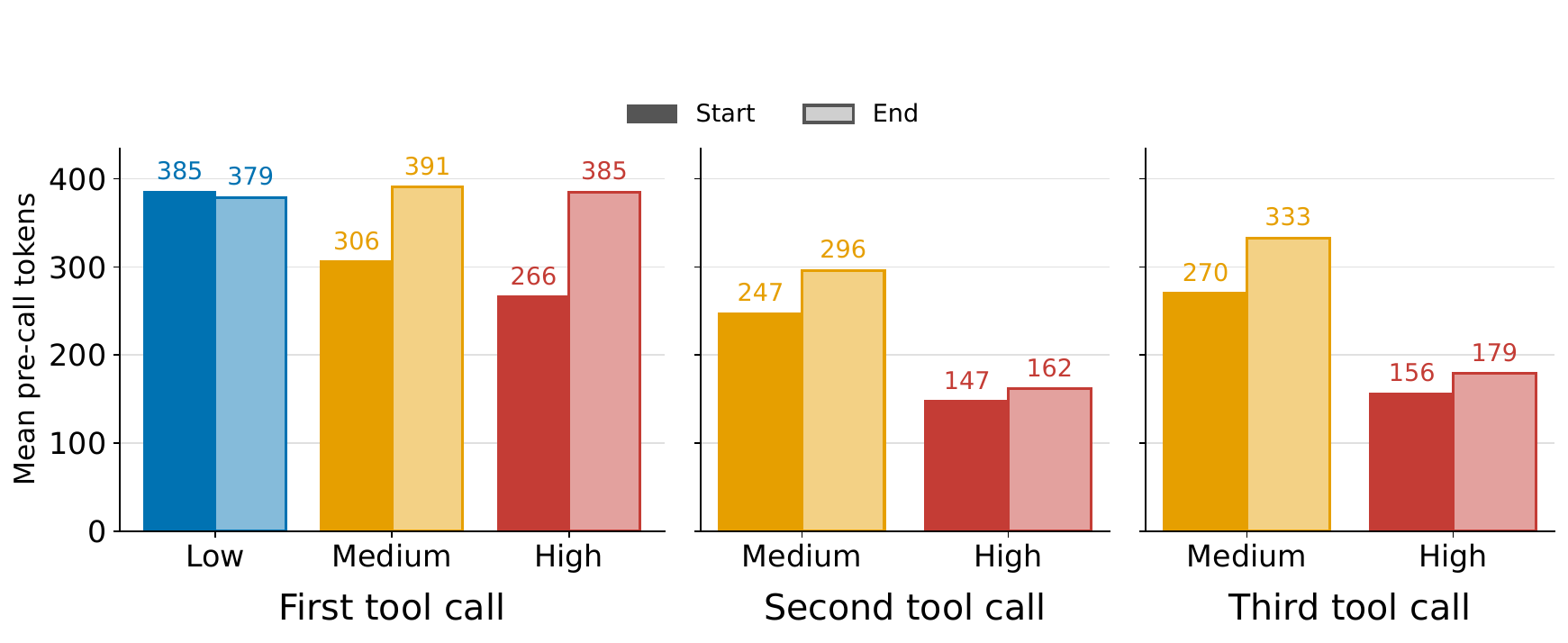}
        \caption{Mean pre-call tokens at RL start and end}
        \label{fig:rl_pre_call_tokens}
    \end{subfigure}
    \caption{
        Pre-call reasoning in SFT and during RL.
        (a) Mean number of reasoning tokens preceding each tool call in the \low, \medium, and \high SFT datasets.
        (b) Mean pre-call reasoning tokens for the first three tool calls at the beginning and end of RL training.
        \high receives substantially less pre-call supervision per call in SFT, while the tool-using trajectories retained by \medium{} and \high{} exhibit longer pre-call reasoning after RL.
    }
    \label{fig:pre_call_tokens}
\end{figure}

Figure~\ref{fig:sft_pre_call_tokens} shows the average number of pre-call tokens, which decreases from 623 tokens in \low{} to 302 in \medium{} and 161 in \high{}. The contrast is particularly pronounced for Code, with means of 991, 423, and 134 tokens, respectively. Thus, \low, \medium, and \high differ not only in tool-call depth but also in the amount of pre-call reasoning. Intuitively, this may cause the calls in \high{} to be insufficiently planned and the information from each tool call to be improperly integrated.

To test whether this hypothesis is consistent with the RL dynamics, we examine whether the tool-using rollouts retained after RL exhibit longer pre-call reasoning. Figure~\ref{fig:rl_pre_call_tokens} compares the start (SFT-trained policy or step=0 of RL training policy) and end (step=32 of RL training policy) of the RL training for all three runs. For both \medium and \high, we observe a consistent increase in the mean pre-call reasoning tokens in the first three calls, which implies that such behaviors tend to produce correct final answers and thus be promoted during the RL training. Specifically, we highlight that this phenomenon is more obvious in the first call, where \medium increases from 306 to 391 and \high increases from 266 to 385. This is natural because the model needs to understand the task, perform basic reasoning and plan the tool calling strategies that lay the foundation for the complete solution. We also note that the difference in \low is very little, mainly because the model already allocates sufficient pre-call tokens that produce robust reasoning ability. Taken together, these observations motivate the following hypothesis:

\begin{tcolorbox}[
    colback=gray!5,
    colframe=black!70,
    title=Hypothesis,
    fonttitle=\small,
    sharp corners
]
\textbf{Pre-call reasoning} is a key factor shaping the \textbf{tool-use calibration}, which in turn affects whether the corresponding tool-use behaviors are retained under RL training.
\end{tcolorbox}

Here, \textbf{tool-use calibration} refers to whether the model invokes tools when they are actually useful, rather than simply learning to call them frequently. Even if a model learns the mechanics of multi-step tool use and sustains high tool-use rates during inference, insufficient pre-call reasoning may lead to poorly calibrated decisions, causing the model to invoke tools unnecessarily.

\subsection{A Theoretical Understanding of Selective Tool Retention}
\label{sec:theory}

Section~\ref{sec:hypothesis} motivates the hypothesis that pre-call reasoning affects the calibration of tool-using trajectories. We examine how different SFT data (\low, \medium, \high) evolve under outcome-only optimization, distinguishing successful tool use from frequent tool invocation.

\paragraph{Modeling Tool Use from SFT to RL.}
Fix a question $x$, and let $\tau$ denote a complete trajectory containing the model's reasoning, tool calls, tool observations, and final answer. Let $p_i(\tau,t)$ denote the trajectory distribution at RL training time $t$ induced by SFT initialization $i$, where different values of $i$ correspond to policies trained on different SFT datasets. For the theoretical analysis, we treat $t$ as continuous optimization time; in the discrete setting, $t$ indexes RL optimization steps. Let $R(\tau)\in\{0,1\}$ denote the outcome reward associated with trajectory $\tau$.
Assume positive initial probability on a common finite trajectory set $\Omega$.
To isolate probability reallocation under outcome-only optimization, we consider the idealized dynamics
\begin{equation}
\frac{\partial p_i(\tau,t)}{\partial t} = \bigl(R(\tau)-A_i(t)\bigr)p_i(\tau,t), \qquad A_i(t)=\mathbb{E}_{p_i(t)}[R(\tau)].
\label{eq:outcome_selection}
\end{equation}
$A_i(t)$ denotes the probability that a rollout ends with a correct answer at time $t$. 
We partition $\Omega$ into the nonempty sets $\mathcal{T}$ and $\mathcal{D}$, containing trajectories with and without at least one supported tool execution attempt, respectively.
Let $I_i(t)$ denote the probability of calling at least one tool in a rollout at time $t$, and let $C_i(t)$ and $D_i(t)$ denote conditional accuracy within $\mathcal{T}$ and $\mathcal{D}$.
Then
\begin{equation}
A_i(t) = I_i(t)C_i(t) + \bigl(1-I_i(t)\bigr)D_i(t).
\label{eq:accuracy_decomposition}
\end{equation}

\begin{theorem}[SFT initialization and selective tool retention]
\label{thm:selective_retention}
Under Eq.~\ref{eq:outcome_selection}, the overall and tool-conditioned accuracy satisfy
\begin{equation}
\dot A_i(t)=A_i(t)\bigl(1-A_i(t)\bigr), \qquad \dot C_i(t)=C_i(t)\bigl(1-C_i(t)\bigr),
\label{eq:accuracy_dynamics}
\end{equation}
whereas tool initiation satisfies
\begin{equation}
\dot I_i(t) = I_i(t)\bigl(1-I_i(t)\bigr) \bigl(C_i(t)-D_i(t)\bigr).
\label{eq:initiation_dynamics}
\end{equation}
Moreover, if two SFT initializations satisfy $A_1(0)\geq A_2(0)$ and $C_1(0)\geq C_2(0)$, then
\begin{equation}
A_1(t)\geq A_2(t), \qquad C_1(t)\geq C_2(t) \qquad\text{for all }t\geq 0.
\label{eq:initialization_ordering}
\end{equation}
These accuracy orderings do not imply $I_1(t)\geq I_2(t)$.
\end{theorem}

Theorem~\ref{thm:selective_retention} distinguishes two effects of SFT initialization: advantages in overall and tool-conditioned accuracy can persist under the same idealized optimization, whereas tool initiation depends on the relative accuracy of the tool-using and tool-free branches. Consequently, tool-conditioned accuracy can improve while initiation decreases whenever $C_i(t)<D_i(t)$.
Demonstrating frequent or deep tool interactions alone, therefore, does not ensure their retention. This motivates selecting SFT trajectories that support reliable pre-call reasoning, rather than preserving tool calls for their own sake. The derivation and proof are provided in Appendix~\ref{app:theory}.

\section{From Diagnosis to SFT Data Design}

In Section~\ref{sec:obs}, we identified pre-call reasoning as a potential factor shaping tool-use calibration. We now ask whether this insight can guide SFT data selection. Specifically, we use average pre-call reasoning length as a proxy and filter the original \high trajectories to test whether this simple intervention improves downstream RL performance.

\subsection{Experimental Setup}
\label{sec:selection_setup}

\paragraph{Data Construction with Pre-call-length-guided Filtering.} 
LF25 and LF50 retain complete trajectories containing approximately 25\% and 50\%, respectively, of the assistant tokens supervised in one epoch of the original \high{} dataset. For a trajectory with $k_i>0$ tool calls, we compute $s_i=\frac{1}{k_i}\sum_{j=1}^{k_i}l_{i,j}$, where $l_{i,j}$ counts the reasoning tokens immediately preceding call $j$. We rank Code and Zoom trajectories separately by $s_i$, taking the highest-ranked examples in each group. Direct-answer trajectories have no pre-call score. We select these separately to meet the token budget while preserving the proportions of Code, Zoom, and direct-answer trajectories in \high{}. Of the original 2,000 trajectories, LF25 retains 320 (16\%) and LF50 retains 720 (36\%).

\paragraph{Training setup.}
We train LF50 for two epochs and LF25 for four epochs, so that their cumulative assistant-supervised token exposure approximately matches Original \high{}. Both models start from the same base checkpoint and retain the original SFT objective and token-masking scheme. The subsequent RL training and evaluations are also identical to those in Section~\ref{sec:obs}.

\subsection{Experimental Results}
\label{sec:filter_exp}

\paragraph{Length filtering improves final task performance.} Table~\ref{tab:length_filtered_results} compares the final RL performance of the \high and length-filtered SFT initializations. The table also reports the ablations discussed in Section~\ref{sec:ablations}. \texttt{LF25} achieves the highest pooled accuracy and leads on four of the five benchmarks among all variants in Table~\ref{tab:length_filtered_results}; \texttt{LF50} outperforms Original \high{} on four of five benchmarks. Thus, concentrating the training budget in SFT on a small subset of the original data pool with longer pre-call reasoning yields better results. This result also resonates with the theoretical findings in Section~\ref{sec:theory}, where we indicate that a stronger initialization can preserve this advantage in task-level success under the same outcome-only optimization.

\begin{table}[h]
    \centering
    \small
    \setlength{\tabcolsep}{5pt}
    \caption{
    Final \texttt{tool-on} accuracy (\%) after RL.
    LF25 and LF50 select tool-using demonstrations by mean pre-call reasoning length. The numbers 25 and 50 denote the percentage of the original \high{} assistant-supervised token budget retained per epoch. The rows below the horizontal rule report the random-selection, total-reasoning-length, and pre-call truncation experiments discussed in Q1--Q3 of Section~\ref{sec:ablations}.
    \underline{Overall}\ is pooled accuracy across all five benchmarks.
    Bold marks column bests. Shading~highlights~LF25.
    }
    \label{tab:length_filtered_results}
    \begin{tabular}{l|ccccc|c}
        \toprule
        Method
        & V*Bench & ChartQA & MathVista & WeMath & CharXiv & \underline{Overall} \\
        \midrule
        \high
        & 84.29 & 85.48 & 65.10 & 57.30 & 31.40 & \underline{66.24} \\
        LF50
        & 84.82 & 85.36 & 67.70 & 58.79 & 37.20 & \underline{67.92} \\
        \rowcolor{bestrow}
        LF25
        & \textbf{86.91} & \textbf{87.72} & \textbf{68.90} & \textbf{60.57} & 38.40 & \underline{\textbf{69.75}} \\
        \midrule
        Random25
        & 81.68 & 82.20 & 66.90 & 60.34 & 35.80 & \underline{66.67} \\
        Matched-Random25 & 85.34 & 87.16 & 67.40 & 57.14 & \textbf{38.80} & 68.39 \\
        Total-Reasoning25 & 76.44 & 83.56 & 66.20 & 54.57 & 34.30 & 65.15 \\
        LF50-Truncated & 75.92 & 84.32 & 63.90 & 58.05 & 33.30 & 65.85 \\
        \bottomrule
    \end{tabular}
\end{table}

\begin{figure}[t]
    \centering
    \includegraphics[width=\linewidth]{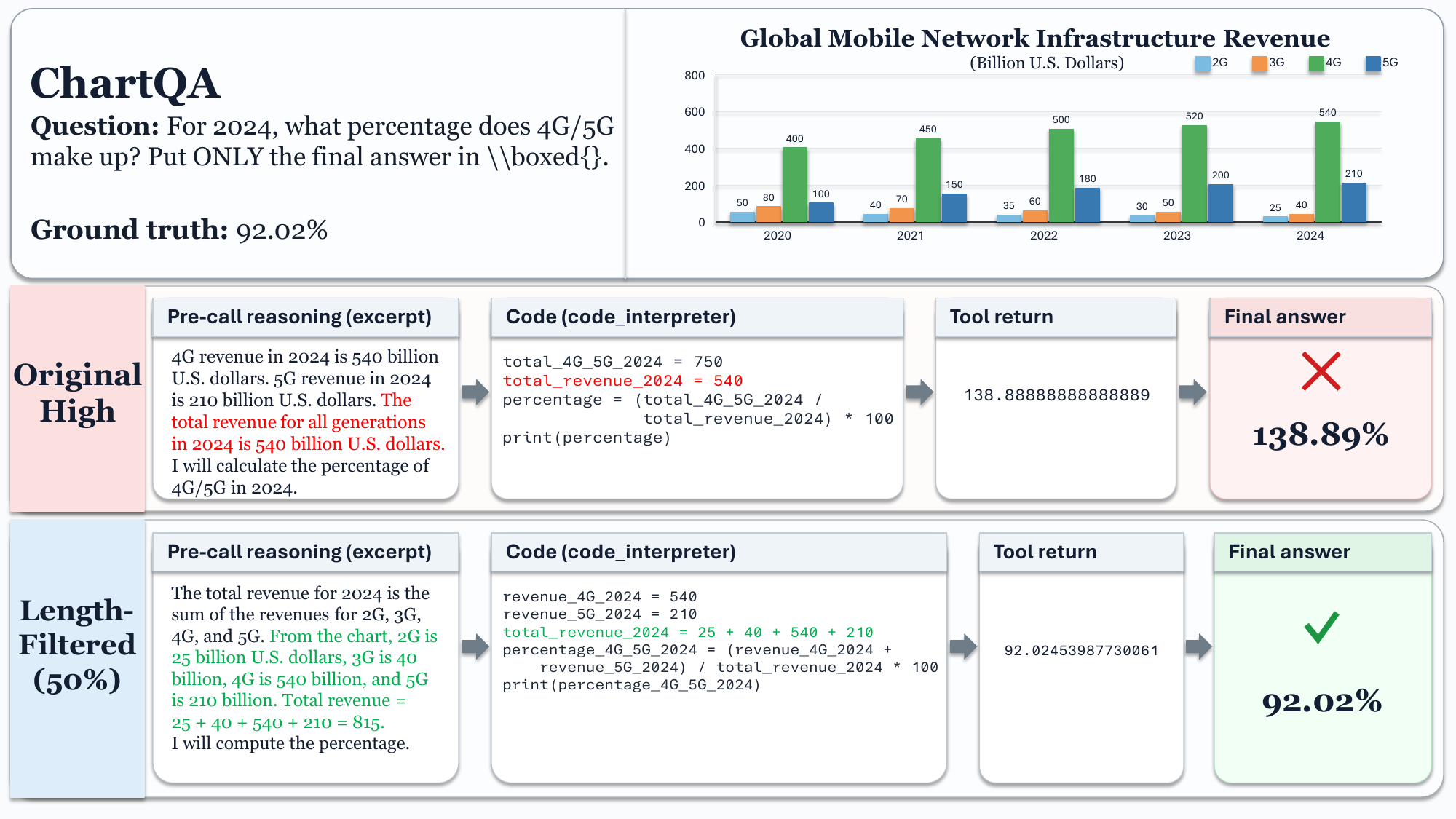}
    \caption{A qualitative ChartQA example illustrating how reasoning before a Code call can determine the computational target. In the original \high trace, the model incorrectly treats the 4G revenue (540) as the total 2024 revenue, leading the subsequent code to use the wrong denominator. In contrast, the Length-Filtered model correctly sums the revenues of all four generations to obtain 815, yielding the correct percentage of 92.02\%.}
    \label{fig:qualitative_example}
\end{figure}

\begin{figure}[t]
    \centering
    \begin{subfigure}[t]{0.48\linewidth}
        \centering
        \includegraphics[width=\linewidth]{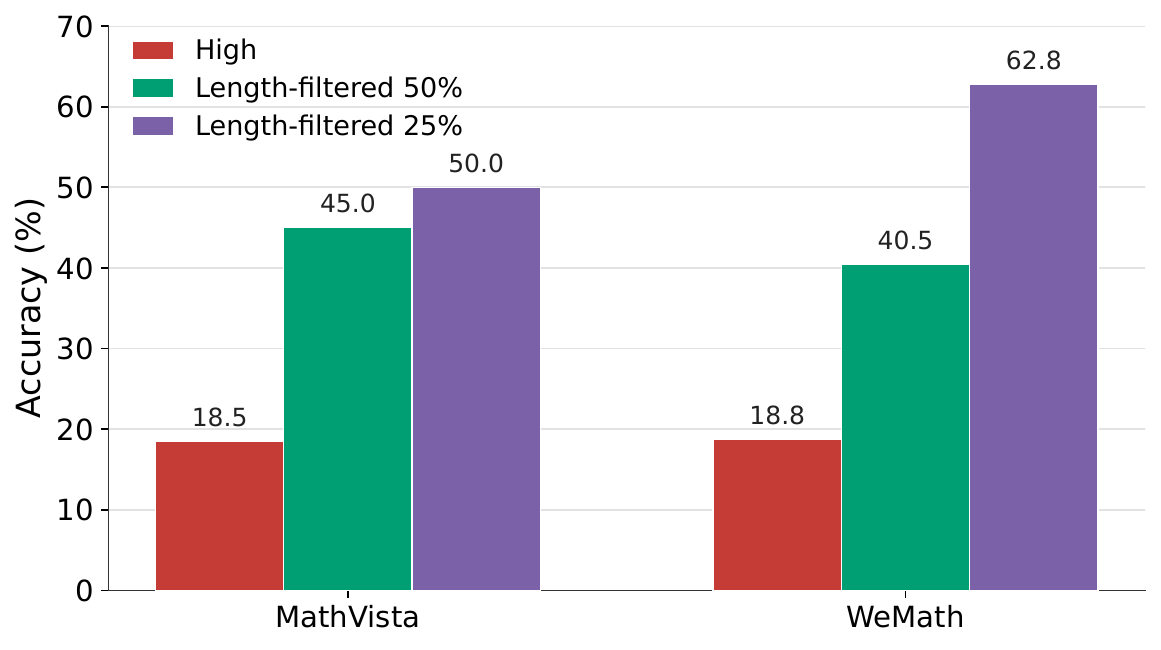}
        \caption{The final accuracy of the tool-using trajectories}
        \label{fig:tool_acc}
    \end{subfigure}
    \hfill
    \begin{subfigure}[t]{0.48\linewidth}
        \centering
        \includegraphics[width=\linewidth]{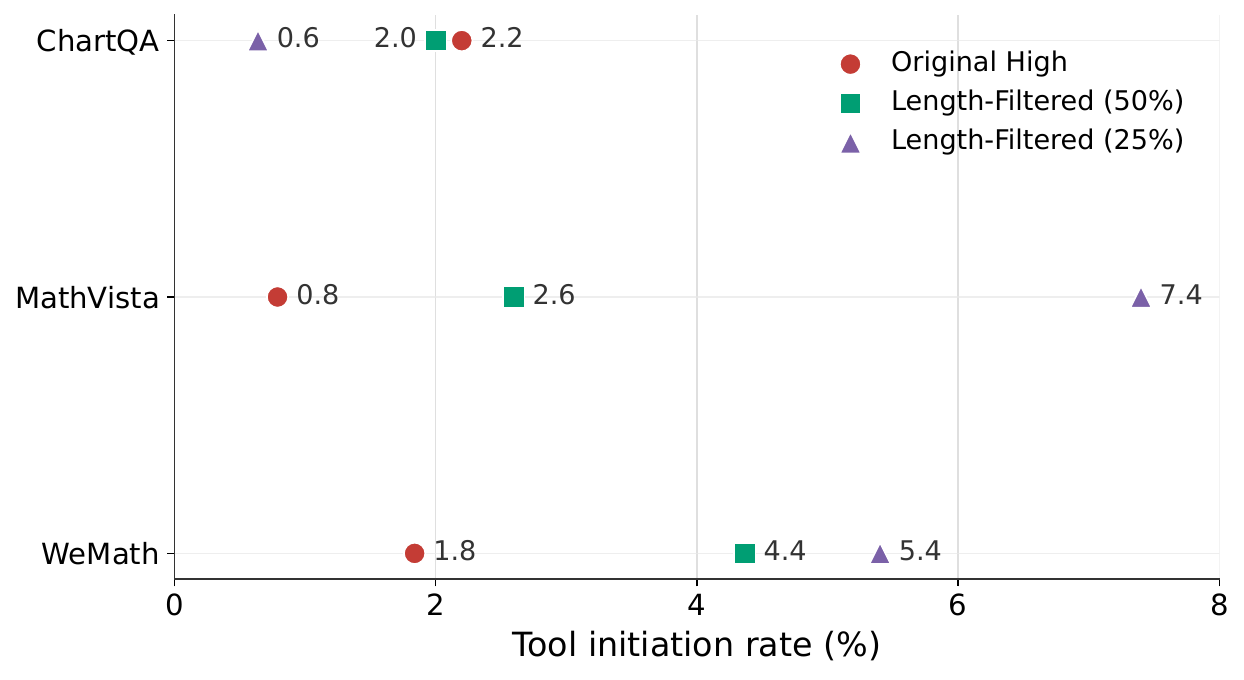}
        \caption{Tool initiation rates after RL}
        \label{fig:tool_init}
    \end{subfigure}
    \caption{Tool-use behavior after RL. Length-filtered SFT yields higher final-answer success among self-selected tool-using trajectories, while tool initiation does not increase uniformly across benchmarks. Together, the two views show that the downstream benefit is better reflected by the effectiveness of retained tool-mediated decisions than by how frequently tools are invoked.}
    \label{fig:tool_acc_init}
\end{figure}

\paragraph{Higher conditional success among tool-using trajectories.} Figure~\ref{fig:tool_acc} reports the final-answer accuracy conditional on trajectories that used tools, indicating that the trajectories on which the filtered policies choose to invoke tools are more likely to end in a correct answer. This improvement is distinct from simply increasing tool usage: it reflects higher success among the model's self-selected tool-involved decisions. Figure~\ref{fig:qualitative_example} provides a qualitative illustration of this distinction: the two traces invoke the same Code tool, but differ in whether the pre-call reasoning identifies the computation that is actually needed to answer the question. This pattern mirrors the ordering of tool-conditioned accuracy in our analysis: a stronger initialization can maintain a higher ratio of successful tool-use trajectories under the same RL training.

\paragraph{Successful completion and tool frequency are distinct.} Figure~\ref{fig:tool_init} compares tool initiation rates across the three settings. Length filtering does not induce a uniform change in how often tools are invoked. On ChartQA, initiation decreases from 2.2\% for Original \high{} to 2.0\% for LF50 and 0.6\% for LF25. In contrast, initiation increases on MathVista and WeMath. Thus, the downstream gains from length filtering cannot be attributed simply to encouraging more tool calls. This behavior follows the tool-initiation dynamics in Eq.~\ref{eq:initiation_dynamics}: tool use is reinforced only when tool-using trajectories outperform the tool-free alternative.

Taken together, these results show that the benefit of length-filtered SFT is not a uniform increase in tool use. Instead, it improves overall task accuracy and the success rate of self-selected tool-using trajectories, while tool initiation varies across benchmarks. This distinction is central to our findings: what matters after RL is not how many tool calls are retained, but whether the retained tool-use decisions are effective for reaching the correct answer.

\subsection{Ablations and Analysis}
\label{sec:ablations}

\paragraph{Q1: Can random selection recover the LF25 gain?}
Random25 tests whether repeating a smaller training set is sufficient. From the same 2,000 \high{} demonstrations, we randomly select whole trajectories without using pre-call length. We match LF25's trajectory counts (120 Code, 120 Zoom, and 80 direct answers) and its assistant-supervised token total within each of the three groups. Together, the 320 trajectories contain approximately 25\% of \high{}'s supervised tokens per epoch, giving Random25 its name. Random25 and LF25 use the same four effective SFT epochs, batch size, 50 optimizer updates, and downstream RL procedure.

Matched-Random25 asks whether LF25 benefits from the kinds of examples its ranking selects. At the same nominal token budget, we draw whole trajectories from \high{} so that the subset approximates LF25's distributions of source dataset, tool type, number of calls per trajectory, and total trajectory length. Selection among trajectories satisfying these composition constraints is random with respect to pre-call length. We then apply the same SFT and RL recipe.

After RL, pooled accuracy is 66.67\% for Random25, 68.39\% for Matched-Random25, and 69.75\% for LF25 (Table~\ref{tab:length_filtered_results}). Random25 trails LF25 on every benchmark despite matching its repetition schedule, tool mix, and token exposure. Matching the broader composition recovers 1.72 percentage points, suggesting that the mix of training examples accounts for part of the observed gap. LF25 still leads by 1.36 points overall and on four of five benchmarks. Thus, pre-call length provides a useful selection signal beyond the measured composition in this comparison. 

\paragraph{Q2: Would selecting the longest reasoning traces work as well?}
Total-Reasoning25 ranks complete \high{} trajectories by the sum of reasoning tokens before all calls and in the final response. LF25 scores tool-using trajectories by mean reasoning length per call. Both subsets target 25\% of the original assistant-supervised tokens per epoch and use the same downstream RL procedure.

As shown in Table~\ref{tab:length_filtered_results}, across all five benchmarks, LF25 scored 69.75\%, compared with 65.15\% for Total-Reasoning25. A trajectory can rank highly on total length simply by making many calls. For example, six calls with 20 reasoning tokens before each contribute more tokens in total than two calls with 50 tokens each. Yet if each of the six calls is only briefly motivated, the model sees the tool invoked repeatedly with little explanation of what information is missing or how the call will obtain it. Averaging over calls removes the advantage of making more calls. LF25's stronger results on all five benchmarks suggest that the reasoning a demonstration provides at each tool decision matters for learning to use tools, and that total length is a poor substitute for this signal.

\paragraph{Q3: What happens when we shorten pre-call reasoning on the same examples?}
Starting with LF50, we edit each pre-call segment in place. We keep the last sentence before the call, then retain the earliest sentences in their original order until about half of the segment's original tokens remain. Only the intervening text is deleted. The questions, tool calls and arguments, observations, and final responses stay the same. We train LF50-Truncated for four SFT epochs and intact LF50 for two to approximately match cumulative exposure to pre-call reasoning tokens. Both models then undergo the same RL procedure.

Shortening the pre-call reasoning reduced accuracy on all five benchmarks, from 67.92\% to 65.85\% overall (Table~\ref{tab:length_filtered_results}). The middle sentences can supply the logical link between what the model already knows and what it needs a tool to find out. For example, a shortened trace may still instruct the model to zoom in on an axis while omitting why the units on that axis determine the answer. Another pass through the same trace repeats the instruction without restoring its rationale. The gap persists at a comparable pre-call token budget, suggesting that the reasoning connecting the evidence to the next action is part of what the demonstration teaches.

\paragraph{Q4: Does tool access help the trained model answer more questions correctly?}
For each post-RL checkpoint (\high{} and LF25), we evaluate the same five benchmark question sets twice with the same decoding settings. With tools enabled, the model may use Zoom or Code or answer directly. With tools disabled, it must answer directly. We compute the difference in accuracy (enabled minus disabled) over all questions, including those on which the model never calls a tool.

With tools disabled, the checkpoints have similar pooled accuracy (68.90\% for \high{}, 68.57\% for LF25). Enabling tools costs \high{} 2.66 percentage points and gives LF25 a 1.18-point gain. The sign of each change holds on all five benchmarks (Table~\ref{tab:tool_availability_gain}). Thus, LF25's advantage emerges when the model can decide whether and how to use tools. This comparison measures the net effect of tool access across questions. To examine which decisions differ, we look next at the calls each policy actually made.

\begin{table}[t]
    \centering
    \small
    \setlength{\tabcolsep}{4pt}
    \caption{
        Tool-availability gain, measured as \texttt{tool-on} minus \texttt{tool-off} accuracy in percentage points. Positive values indicate higher accuracy with tools available.
    }
    \label{tab:tool_availability_gain}
    \begin{tabular}{lrrrrrr}
        \toprule
        Initialization
        & V*Bench & ChartQA & MathVista & WeMath & CharXiv & Overall \\
        \midrule
        Original \high{}
        & -1.57 & -2.08 & -1.80 & -3.68 & -3.40 & -2.66 \\
        \rowcolor{bestrow}
        LF25
        & \textbf{+5.23} & \textbf{+1.00} & \textbf{+1.30} & \textbf{+1.37} & \textbf{+0.40} & \textbf{+1.18} \\
        \bottomrule
    \end{tabular}
\end{table}

\input{behavior_assessment}

\FloatBarrier
\begin{tcolorbox}[
    colback=gray!5,
    colframe=black!70,
    title={Takeaways from the ablations},
    fonttitle=\small,
    sharp corners
]
\textbf{1. The quality of pre-call reasoning matters (Q1--Q3).}
Even after approximately matching LF25's distributions of data source, tool type, tool-call depth, and trajectory length, random selection still falls short of LF25 in pooled accuracy. Total reasoning length is also a poor substitute for average pre-call reasoning length. A demonstration can accumulate many reasoning tokens across several calls while giving little justification for any individual call. The truncation experiment provides more direct evidence. Removing intermediate reasoning steps from the same demonstrations lowers post-RL accuracy even at a comparable cumulative pre-call token budget. Together, these results suggest that an important part of what the model learns comes from the reasoning that connects the available evidence to the next tool call.

\smallskip
\textbf{2. Filtering helps the model use tools more effectively (Q4--Q5).}
After RL, enabling tools improves LF25's accuracy on all five benchmarks, with a pooled gain of 1.18 percentage points. On both MathVista and WeMath, LF25 outperforms \high{} on two aspects of tool use. It more often chooses a suitable tool and supplies appropriate arguments for the same call. It also more often interprets a successful Code result correctly and uses it in the subsequent reasoning or answer. Both advantages are visible immediately after SFT and persist after RL.
\end{tcolorbox}

\section{Conclusion}

We study how SFT shapes the tool-use behaviors retained under outcome-only RL. With approximately matched SFT budgets but different tool-call depths, we observe similar RL training accuracy yet sharply different tool-use dynamics, motivating pre-call reasoning as a key factor in tool-use calibration. Our selective-retention analysis shows that advantages in overall and tool-conditioned accuracy can persist under the same optimization, while tool initiation depends on whether tool use is more successful than the tool-free alternative. Guided by this view, pre-call-reasoning-length-guided filtering improves post-RL accuracy and tool-enabled performance across five benchmarks, outperforming a matched random-selection control without consistently increasing tool initiation. Overall, effective TIR training should prioritize supervision that supports well-calibrated tool use rather than tool-call frequency alone. 

A natural future direction is to move beyond fixed proxies such as pre-call reasoning length and develop more adaptive criteria for selecting or generating SFT data that better prepares tool use for downstream RL. More broadly, it would be interesting to automate this process by jointly selecting, rewriting, or generating SFT trajectories based on downstream RL behavior, rather than relying on a manually chosen filtering rule.

\section*{Acknowledgments}
This research was supported by the U.S. Army Research Laboratory and the U.S. Army Research Office under Grant W911NF2010219, the Office of Naval Research, the National Science Foundation, and Meta. Joseph Suh acknowledges support from the Korea Foundation for Advanced Studies through the Doctoral Study Abroad Program.

\bibliographystyle{assets/plainnat}
\bibliography{references}

\clearpage
\appendix

\section{Omitted Proof}
\label{app:theory}

We condition throughout on a fixed question $x$ and use the notation
of Section~\ref{sec:theory}.
The finite trajectory set $\Omega$ and terminal reward
$R:\Omega\rightarrow\{0,1\}$ are fixed.
Both initializations assign positive probability to every
trajectory in $\Omega$, and both $\mathcal{T}$ and $\mathcal{D}$
are nonempty.

\subsection{Derivation of Eq.~\ref{eq:outcome_selection}}
\label{app:outcome_dynamics}

We derive Eq.~\ref{eq:outcome_selection} from an idealized KL-proximal update over trajectory distributions. This update is used only to isolate reward-driven probability reallocation; it is not an exact reformulation of the GRPO objective or its reference-policy KL regularization.

For clarity, we omit the initialization index $i$. Let $p(\tau,t)$ denote the probability assigned to trajectory $\tau$ at optimization time $t$, and let $R(\tau)\in\{0,1\}$ denote its fixed terminal correctness reward.

For an infinitesimal update interval $h>0$, let $q(\tau)$ denote the candidate trajectory distribution for the next update and consider
\begin{equation}
q^\star = \operatorname*{arg\,max}_{q\in\Delta(\Omega)} \left\{\sum_{\tau\in\Omega}q(\tau)R(\tau) - \frac{1}{h} D_{\mathrm{KL}}\bigl(q\|p(\cdot,t)\bigr)\right\},
\label{eq:proximal_update}
\end{equation}
where $\Delta(\Omega)$ is the probability simplex.
We then set
\[
p(\tau,t+h)=q^\star(\tau).
\]

Using
\[
D_{\mathrm{KL}}\bigl(q\|p(\cdot,t)\bigr) = \sum_{\tau\in\Omega}q(\tau)\log\frac{q(\tau)}{p(\tau,t)},
\]
and introducing a Lagrange multiplier $\alpha$ for the constraint
$\sum_\tau q(\tau)=1$, we obtain
\[
\mathcal{L}(q,\alpha) = \sum_\tau q(\tau)R(\tau) - \frac{1}{h} \sum_\tau q(\tau)\log\frac{q(\tau)}{p(\tau,t)} + \alpha\left(\sum_\tau q(\tau)-1\right).
\]

Differentiating with respect to $q(\tau)$ gives
\[
R(\tau) - \frac{1}{h}\left(\log\frac{q(\tau)}{p(\tau,t)}+1\right) + \alpha = 0.
\]

Rearranging,
\[
\log\frac{q(\tau)}{p(\tau,t)} = hR(\tau)+h\alpha-1.
\]
The last two terms are independent of $\tau$ and can therefore be absorbed into the normalization constant. Hence,
\begin{equation}
p(\tau,t+h) = \frac{p(\tau,t)e^{hR(\tau)}}{Z(t,h)},
\label{eq:exponential_reweighting}
\end{equation}
where
\[
Z(t,h) = \sum_{\tau'} p(\tau',t)e^{hR(\tau')}.
\]

For $h\rightarrow0$,
\[
e^{hR(\tau)} = 1+hR(\tau)+o(h).
\]
Therefore,
\[
\begin{aligned}
Z(t,h) &= \sum_{\tau'} p(\tau',t) \left(1+hR(\tau')+o(h)\right)\\
&= 1+h\sum_{\tau'}p(\tau',t)R(\tau')+o(h)\\
&= 1+hA(t)+o(h),
\end{aligned}
\]
where
\[
A(t) = \sum_{\tau'}p(\tau',t)R(\tau')
\]
is the expected final-answer accuracy.

Substituting this expansion into Eq.~\ref{eq:exponential_reweighting},
\[
p(\tau,t+h) = p(\tau,t)\frac{1+hR(\tau)+o(h)}{1+hA(t)+o(h)}.
\]
Thus,
\[
\frac{p(\tau,t+h)-p(\tau,t)}{h} = p(\tau, t)\frac{R(\tau)-A(t)+o(1)}{1+hA(t)+o(h)}.
\]
Taking $h\rightarrow0$ gives
\[
\frac{\partial p(\tau,t)}{\partial t} = \bigl(R(\tau)-A(t)\bigr)p(\tau,t).
\]

\subsection{Proof of Theorem~\ref{thm:selective_retention}}
\label{app:selective_retention_proof}

\begin{proof}
We first derive the dynamics of overall accuracy,
tool-conditioned accuracy, and tool initiation, and then show
that the accuracy ordering induced by the SFT initialization
is preserved.

\paragraph{Overall accuracy.}
By definition,
\[
A_i(t) = \sum_{\tau\in\Omega} p_i(\tau,t)R(\tau).
\]
Since $R(\tau)$ is fixed,
\[
\begin{aligned}
\dot A_i(t) &= \sum_{\tau} R(\tau)\frac{\partial p_i(\tau,t)}{\partial t}\\
&= \sum_{\tau} p_i(\tau,t)R(\tau) \bigl(R(\tau)-A_i(t)\bigr)\\
&= \sum_{\tau} p_i(\tau,t)R(\tau)^2 - A_i(t) \sum_{\tau} p_i(\tau,t)R(\tau).
\end{aligned}
\]
Because $R(\tau)\in\{0,1\}$, we have $R(\tau)^2=R(\tau)$. Both sums therefore equal $A_i(t)$, giving
\[
\dot A_i(t) = A_i(t)\bigl(1-A_i(t)\bigr).
\]

\paragraph{Tool-initiation dynamics.}
Define the probability mass of correct tool-using trajectories as
\[
N_i(t) := \sum_{\tau\in\mathcal{T}} p_i(\tau,t)R(\tau).
\]
By the definition of conditional accuracy,
\[
N_i(t) = I_i(t)C_i(t).
\]

Since
\[
I_i(t) = \sum_{\tau\in\mathcal{T}} p_i(\tau,t),
\]
Eq.~\ref{eq:outcome_selection} gives
\[
\begin{aligned}
\dot I_i(t) &= \sum_{\tau\in\mathcal{T}}\frac{\partial p_i(\tau,t)}{\partial t}\\
&= \sum_{\tau\in\mathcal{T}}p_i(\tau,t)\bigl(R(\tau)-A_i(t)\bigr)\\
&= N_i(t)-A_i(t)I_i(t)\\
&= I_i(t)\bigl(C_i(t)-A_i(t)\bigr).
\end{aligned}
\]
Using
\[
A_i(t) = I_i(t)C_i(t) + \bigl(1-I_i(t)\bigr)D_i(t),
\]
we obtain
\[
\begin{aligned}
C_i(t)-A_i(t) &= C_i(t) - I_i(t)C_i(t) - \bigl(1-I_i(t)\bigr)D_i(t)\\
&= \bigl(1-I_i(t)\bigr) \bigl(C_i(t)-D_i(t)\bigr).
\end{aligned}
\]
Therefore,
\[
\dot I_i(t) = I_i(t)\bigl(1-I_i(t)\bigr) \bigl(C_i(t)-D_i(t)\bigr).
\]

\paragraph{Tool-conditioned accuracy.}
Differentiating $N_i(t)$ gives
\[
\begin{aligned}
\dot N_i(t) &= \sum_{\tau\in\mathcal{T}} R(\tau) \frac{\partial p_i(\tau,t)}{\partial t}\\
&= \sum_{\tau\in\mathcal{T}} p_i(\tau,t)R(\tau) \bigl(R(\tau)-A_i(t)\bigr).
\end{aligned}
\]
Again using $R(\tau)^2=R(\tau)$,
\[
\begin{aligned}
\dot N_i(t) &= \sum_{\tau\in\mathcal{T}} p_i(\tau,t)R(\tau) - A_i(t) \sum_{\tau\in\mathcal{T}} p_i(\tau,t)R(\tau)\\
&= N_i(t)\bigl(1-A_i(t)\bigr).
\end{aligned}
\]

Since
\[
C_i(t) = \frac{N_i(t)}{I_i(t)},
\]
we have
\[
\dot C_i(t) = \frac{\dot N_i(t)I_i(t) - N_i(t)\dot I_i(t)}{I_i(t)^2}.
\]
Substituting the expressions above,
\[
\begin{aligned}
\dot C_i(t) &= \frac{N_i(t)(1-A_i(t))I_i(t) - N_i(t)I_i(t)(C_i(t)-A_i(t))}{I_i(t)^2}\\
&= \frac{N_i(t)}{I_i(t)} \left[(1-A_i(t))-(C_i(t)-A_i(t))\right]\\
&= C_i(t)\bigl(1-C_i(t)\bigr).
\end{aligned}
\]

\paragraph{Preservation of initialization ordering.}
From the previous derivation, we know that both $A_i(t)$ and $C_i(t)$ satisfy the scalar differential equation
\[
\dot y(t) = y(t)(1-y(t)).
\]
Consider first an initial value $y(0)=y_0\in(0,1)$.
Separating the variables gives
\[
\frac{dy}{y(1-y)}=dt.
\]
Integrating the left-hand side with respect to $y$ from
$y_0$ to $y(t)$, and the right-hand side with respect to time
from $0$ to $t$, we obtain
\[
\int_{y_0}^{y(t)}\frac{1}{y(1-y)}\,dy = \int_0^t ds.
\]
The left-hand side is
\[
\begin{aligned}
\int_{y_0}^{y(t)}\frac{1}{y(1-y)}\,dy &=
\int_{y_0}^{y(t)} \left(\frac{1}{y} + \frac{1}{1-y} \right)dy\\
&= \left[\log y-\log(1-y)\right]\Bigg|_{y_0}^{y(t)}\\
&= \log\frac{y(t)}{1-y(t)} - \log\frac{y_0}{1-y_0}.
\end{aligned}
\]
Therefore,
\[
\log\frac{y(t)}{1-y(t)} - \log\frac{y_0}{1-y_0} = t.
\]
Exponentiating both sides gives
\[
\frac{y(t)}{1-y(t)} = \frac{y_0}{1-y_0}e^t.
\]
Solving for $y(t)$ yields
\[
y(t) = \frac{y_0e^t}{1-y_0+y_0e^t}.
\]

Define
\[
F_t(u) = \frac{ue^t}{1-u+ue^t}.
\]
For every $t\geq0$, we have
\[
\frac{\partial F_t(u)}{\partial u} = \frac{e^t}{\left(1-u+ue^t\right)^2} > 0.
\]
Therefore, $F_t$ is increasing in its initial value.
It follows that
\[
A_1(0) \geq A_2(0) \quad\Longrightarrow\quad A_1(t) \geq A_2(t),
\]
and similarly,
\[
C_1(0) \geq C_2(0) \quad\Longrightarrow\quad C_1(t) \geq C_2(t)
\]
for all $t\geq0$.
\end{proof}

\section{Prompts}
  \label{app:eval_prompts}

  \begin{promptbox}{TOOL-OFF PROMPT}
  <|im_start|>system
  You are a careful multimodal reasoner. Solve the problem from the image and question.
  <|im_end|>
  <|im_start|>user
  <image>
  {QUESTION}
  {BENCHMARK-SPECIFIC ANSWER-FORMAT INSTRUCTION}
  <|im_end|>
  <|im_start|>assistant
  \end{promptbox}

  \begin{promptbox}{TOOL-ON PROMPT}
  <|im_start|>system
  You are a careful multimodal reasoner. Solve the problem from the image and question. If tools are available, use them only
  when they are genuinely helpful. Use image_zoom_in_tool only for closer visual inspection. Use code_interpreter only for
  genuine numerical or symbolic computation over quantities and relations you have already extracted; it cannot inspect the
  original image.

  # Tools

  You may call one or more functions to assist with the user query.
  Function signatures are provided within <tools></tools> XML tags:

  <tools>
  {
    "function": {
      "description": "Crop and magnify a rectangular region of the input image for closer visual inspection.",
      "name": "image_zoom_in_tool",
      "parameters": {
        "properties": {
          "bbox_2d": {
            "type": "array",
            "description": "Bounding box [x1, y1, x2, y2] in original-image pixel coordinates."
          },
          "code": null,
          "label": {
            "type": "string",
            "description": "Optional label for the region."
          }
        },
        "required": ["bbox_2d"],
        "type": "object"
      }
    },
    "type": "function"
  }
  {
    "function": {
      "description": "Execute Python for genuine computation over quantities and relations already extracted from the
      problem; the original image is not an input to this tool.",
      "name": "code_interpreter",
      "parameters": {
        "properties": {
          "bbox_2d": null,
          "code": {
            "type": "string",
            "description": "Python source for numerical or symbolic computation."
          },
          "label": null
        },
        "required": ["code"],
        "type": "object"
      }
    },
    "type": "function"
  }
  </tools>

  Return each function call as JSON within <tool_call></tool_call> XML tags:

  <tool_call>
  {"name": <function-name>, "arguments": <args-json-object>}
  </tool_call>
  <|im_end|>
  <|im_start|>user
  <image>
  {QUESTION}
  {BENCHMARK-SPECIFIC ANSWER-FORMAT INSTRUCTION}
  <|im_end|>
  <|im_start|>assistant
  \end{promptbox}

\section{Experimental Configuration}
\label{app:exp_config}

\subsection{SFT Configuration}
\label{app:sft_config}

The main experiments initialize from Qwen2.5-VL-7B-Instruct. Appendix~\ref{app:qwen35_backbone} describes the Qwen3.5-4B comparison. For the original \low, \medium, and \high settings, each training set contains 2,000 question--trajectory pairs containing 750 Zoom trajectories, 750 Code trajectories, and 500 shared tool-free ones. \low contains one-call tool trajectories, \medium contains two-to-three-call trajectories, and \high contains trajectories with at least four calls. The three datasets have approximately matched total token budgets to ensure a fair comparison on the SFT training stage.

We train each model for one epoch, corresponding to 50 optimizer updates, using AdamW with a global batch size of 40. The learning rate peaks at $1\times10^{-5}$ and is cosine-decayed to $1\times10^{-6}$. The SFT objective is computed only over assistant-generated tokens, including natural-language reasoning, structured tool calls, and final answers. System messages, user messages, and tool observations are masked from the loss. All three settings use the same prompt templates, tool schemas, optimizer configuration, and training schedule.

For the length-filtering experiments in Section~\ref{sec:selection_setup}, we retain complete trajectories from the original \high dataset without modifying their contents. Length-Filtered (50\%) and Length-Filtered (25\%) retain approximately 50\% and 25\% of the original assistant-supervised token mass per epoch, corresponding to approximately 36\% and 16\% of the trajectories, respectively.
To approximately match the cumulative assistant-supervised token exposure of the original \high while keeping the optimization budget fixed, all variants use exactly 50 optimizer updates. The original \high is trained for one epoch with a global batch size of 40, whereas Length-Filtered (50\%) and Length-Filtered (25\%) are trained for two and four epochs with global batch sizes of 29 and 26, respectively. The optimizer, learning-rate schedule, warmup, weight decay, loss masking, and all other SFT settings are kept unchanged.

\subsection{RL Configuration}
\label{app:rl_config}

Starting from each SFT checkpoint, we apply the same outcome-only GRPO~\citep{shao2024deepseekmath} procedure. The main runs use 32 RL updates (64 in Appendix~\ref{app:longer_rl}) with a learning rate of $1\times10^{-6}$. At each update, we sample 512 training questions and generate five rollouts per question. All model variants use the same RL questions, question ordering, sampling configuration, and decoding settings.

Each rollout may invoke Zoom and Code for at most eight tool calls and has an 8,192-token response budget. The reward is determined solely by final-answer correctness. We do not provide auxiliary rewards for tool use, tool-call depth, pre-call reasoning length, or intermediate reasoning quality. Consequently, any change in tool-use behavior during RL arises from optimizing final task success rather than from explicit supervision of tool frequency.

\subsection{Evaluation Configuration}
\label{app:eval_config}

We report two inference settings. In \texttt{tool-on}, both Zoom and Code are available, with the same maximum of eight tool calls and 8,192-token response budget used during RL. The model decides itself whether to call tools and when to call them. It may invoke multiple tools or simply perform a tool-free rollout based on its own decision. In \texttt{tool-off}, the tool schemas are removed and the model must answer directly. Final accuracy is computed from the model's final answer under the benchmark-specific evaluation protocol.

For tool-behavior analysis, we define a \emph{tool-using trajectory} as one containing at least one actual execution attempt of a supported Zoom or Code call. The \emph{tool initiation rate} is the fraction of evaluation examples satisfying this criterion. We further report \emph{tool-conditioned accuracy}, defined as final-answer accuracy restricted to these tool-using trajectories. Because different policies may choose to invoke tools on different questions, this conditional metric is descriptive and should not be interpreted as a same-question causal estimate of the benefit of tool use. Malformed calls or unknown tool names that do not result in an execution are not counted as successful tool initiations.

\section{Additional Results}

Tables~\ref{tab:tool_on_full} and~\ref{tab:tool_off_full} provide the benchmark-level breakdown of the aggregate results reported in Table~\ref{tab:eval}. RL improves final accuracy broadly across benchmarks for all three SFT initializations, despite their substantially different tool-use dynamics discussed in Section~\ref{sec:obs}. The detailed results also show that the aggregate trends are not driven by a single benchmark: after RL, the three initializations achieve broadly comparable \texttt{tool-on} performance on several benchmarks, while \texttt{tool-off} performance remains competitive.

\begin{table*}[h]
    \centering
    \caption{
        Tool-on accuracy (\%) of the \low, \medium, and \high SFT and RL
        checkpoints. Overall reports pooled accuracy across all 6,431
        evaluation examples.
    }
    \label{tab:tool_on_full}
    \small
    \setlength{\tabcolsep}{9pt}
    \begin{tabular}{lrrrrrr}
        \toprule
        \textbf{Checkpoint}
        & \textbf{V*}
        & \textbf{ChartQA}
        & \textbf{MathVista}
        & \textbf{WeMath}
        & \textbf{CharXiv}
        & \textbf{Overall} \\
        \midrule
        \low{} SFT
        & 79.06 & 71.80 & 57.90 & 54.89 & 33.60 & 59.34 \\
        \low{} RL
        & 83.25 & 84.72 & 65.60 & 58.10 & 40.80 & 67.67 \\
        \midrule
        \medium{} SFT
        & 54.97 & 53.96 & 48.80 & 53.05 & 18.50 & 47.43 \\
        \medium{} RL
        & 85.34 & 83.40 & 66.90 & 62.24 & 35.10 & 67.66 \\
        \midrule
        \high{} SFT
        & 26.70 & 48.00 & 47.30 & 55.57 & 10.30 & 43.45 \\
        \high{} RL
        & 84.29 & 85.48 & 65.10 & 57.30 & 31.40 & 66.24 \\
        \bottomrule
    \end{tabular}
\end{table*}

\begin{table*}[h]
    \centering
    \caption{
        Tool-off accuracy (\%) of the \low, \medium, and \high SFT and RL checkpoints. Overall reports pooled accuracy across all 6,431 evaluation examples.
    }
    \label{tab:tool_off_full}
    \small
    \setlength{\tabcolsep}{9pt}
    \begin{tabular}{lrrrrrr}
        \toprule
        \textbf{Checkpoint}
        & \textbf{V*}
        & \textbf{ChartQA}
        & \textbf{MathVista}
        & \textbf{WeMath}
        & \textbf{CharXiv}
        & \textbf{Overall} \\
        \midrule
        \low{} SFT
        & 75.39 & 74.04 & 61.20 & 55.17 & 26.90 & 59.65 \\
        \low{} RL
        & 83.77 & 86.16 & 69.00 & 56.72 & 36.00 & 67.66 \\
        \midrule
        \medium{} SFT
        & 80.63 & 76.44 & 62.60 & 55.80 & 33.40 & 62.14 \\
        \medium{} RL
        & 85.34 & 87.88 & 68.50 & 62.93 & 39.80 & 70.56 \\
        \midrule
        \high{} SFT
        & 79.58 & 78.20 & 65.10 & 55.46 & 36.20 & 63.52 \\
        \high{} RL
        & 85.86 & 87.56 & 66.90 & 60.98 & 34.80 & 68.90 \\
        \bottomrule
    \end{tabular}
\end{table*}

\input{supplementary_experiments}

\end{document}

%% file: behavior_assessment.tex
\paragraph{Q5: Which parts of tool use improve with LF25?}
We use GPT-6.1-Sol as the judge. On MathVista and WeMath, we randomly sample tool-using trajectories from the \high{} and LF25 models, both immediately after SFT and after RL. The judge assesses three decisions (Table~\ref{tab:rebuttal_behavior}).
\begin{enumerate}
\setlength{\itemsep}{2pt}\setlength{\parskip}{0pt}
\item \textbf{First-call justification.} Given the question and what the model knew just before its first call, was there a reason to use a tool? For example, a call can be justified if the model still needs to read a small label in an image. Each trajectory contributes one decision.
\item \textbf{Appropriate tool and arguments.} Given the evidence available before each call, did the model choose a suitable tool and supply arguments that carry out the intended operation? Both conditions must hold. A Zoom call aimed at the wrong region or a Code call that computes a ratio with the wrong denominator fails this criterion.
\item \textbf{Correct use of a successful Code output.} When Code returned a result, did the model interpret it correctly and use it in the next steps or final answer? For example, quoting a computed ratio but applying it to the wrong quantity would fail. The denominator is all successful Code calls in the sample.
\end{enumerate}

The clearest LF25 gain is in \emph{using the result that Code returns}. After RL, correct use rises from 41.9\% to 71.3\% on MathVista and from 58.5\% to 84.2\% on WeMath. The share of calls with both an appropriate tool and appropriate arguments also rises, from 21.5\% to 35.5\% on MathVista and from 57.1\% to 65.6\% on WeMath. Both advantages are already visible after SFT. By comparison, the judge favors \high{} on MathVista and LF25 on WeMath for the decision to make the first call after RL. In these sampled interactions, filtering helps most consistently with constructing an appropriate call and using the evidence it returns. Appendix~\ref{app:rebuttal_behavior_protocol} gives the sampling and judging protocol.

\begin{table}[htbp]
    \centering
    \small
    \setlength{\tabcolsep}{4pt}
    \renewcommand{\arraystretch}{1.32}
    \caption{GPT-6.1-Sol behavior assessments (\%). We assess justification on first calls, joint tool and argument appropriateness on all observed calls, and correct output use on successful Code calls. Bold marks the higher rate within each benchmark and stage.}
    \label{tab:rebuttal_behavior}
    \begin{tabularx}{\linewidth}{@{}lccC{1}C{1}C{1}@{}}
    \toprule
    Benchmark & Stage & Init. & First-call justification & Appropriate tool and arguments & Correct use of successful Code outputs \\
    \midrule
    \multirow{4}{*}{MathVista}
      & \multirow{2}{*}{SFT} & \high{} & 31.2 & 33.3 & 7.7 \\
      & & LF25 & \textbf{34.4} & \textbf{34.9} & \textbf{16.0} \\
    \cmidrule(lr){2-6}
      & \multirow{2}{*}{RL} & \high{} & \textbf{62.5} & 21.5 & 41.9 \\
      & & LF25 & 46.9 & \textbf{35.5} & \textbf{71.3} \\
    \midrule
    \multirow{4}{*}{WeMath}
      & \multirow{2}{*}{SFT} & \high{} & 28.1 & 37.6 & 64.3 \\
      & & LF25 & \textbf{46.9} & \textbf{42.9} & \textbf{72.4} \\
    \cmidrule(lr){2-6}
      & \multirow{2}{*}{RL} & \high{} & 18.2 & 57.1 & 58.5 \\
      & & LF25 & \textbf{43.8} & \textbf{65.6} & \textbf{84.2} \\
    \bottomrule
    \end{tabularx}
\end{table}

%% file: supplementary_experiments.tex
\begingroup
\section{Ablations of Data Selection and Tool Use}
\label{app:rebuttal}\label{app:rebuttal_controls}
Section~\ref{sec:ablations} asks five questions about length-filtered SFT. Q1 uses two random-selection controls, each described below. Q2 changes the selection rule, Q3 edits fixed demonstrations, and Q4--Q5 examine the trained models at inference. All experiments in this section use Qwen2.5-VL-7B-Instruct. The RL stage uses 32 updates of outcome-only GRPO as described in Appendix~\ref{app:rl_config}.

\subsection{Random Trajectory Selection (Q1)}
\label{app:random_selection}
We construct Random25 from the same 2,000 \high{} demonstrations used for LF25. We select complete trajectories separately from the Code, Zoom, and direct-answer groups. Within each group, we assign each candidate a fixed-seed random weight and select a subset that matches LF25 exactly on three quantities. These are the number of trajectories, the sum of assistant-supervised tokens recorded in the dataset, and the sum of assistant tokens that receive loss after applying the chat template. This gives 120 Code, 120 Zoom, and 80 direct-answer trajectories, with approximately one quarter of the original \high{} supervision per pass. A trajectory's pre-call score plays no role in the draw. Both Random25 and LF25 use four effective passes, a global SFT batch size of 26, and 50 optimizer updates. Repeated examples used to fill the final batches are accounted for in the cumulative token match. Table~\ref{tab:length_filtered_results} compares the resulting models after the same RL training.

\subsection{Composition-Matched Random Selection (Q1)}
\label{app:matched_selection}
Matched-Random25 asks whether the LF25 selection rule favors a particular mix of demonstrations. We sample whole trajectories from \high{} at the nominal 25\% supervised-token budget. The target is LF25's composition along four dimensions. We match the dataset each trajectory comes from, whether it uses Zoom or Code or answers directly, the number of calls in the trajectory, and its total length. These distributions are matched approximately. Within those constraints the draw does not rank candidates by pre-call length. We apply the same SFT and downstream RL recipe and compare pooled tool-on accuracy with both Random25 and LF25. The three-way comparison in Table~\ref{tab:length_filtered_results} shows how much of the observed gap narrows when the measured composition is matched.

\subsection{Selection by Total Reasoning Length (Q2)}
\label{app:total_reasoning}
For each candidate in \high{}, Total-Reasoning25 counts the natural-language reasoning tokens before all calls and in the final response. It ranks whole demonstrations by this sum. Tool instructions, their arguments, and returned observations do not enter the reasoning score. LF25 instead scores a tool-using demonstration by the mean number of reasoning tokens before a call (Section~\ref{sec:selection_setup}). For example, two demonstrations with equally long pre-call segments can receive different total-length scores if one makes more calls or gives a longer final explanation. Both methods retain complete trajectories under the same nominal budget of 25\% of \high{}'s assistant-supervised tokens per epoch. We compare the post-RL checkpoints with tools enabled on the five benchmarks in Table~\ref{tab:length_filtered_results}.

\subsection{Truncation of Pre-Call Reasoning (Q3)}
\label{app:precall_truncation}
We begin with the 720 LF50 trajectories. Within each tool-using demonstration we edit each pre-call segment independently. We split it into sentences, reserve the final sentence immediately before the call, and keep the earliest sentences in order until the retained text is approximately half of the original segment's token count. The gap between the prefix and the final sentence is deleted without any generated replacement. The target is approximate because whole sentences are retained. Direct-answer trajectories have no such segment to edit. We preserve the questions, tool names and arguments, observations, and final reasoning and answers. Intact LF50 receives two effective SFT epochs and LF50-Truncated receives four. With about half the pre-call tokens retained on each pass, this schedule approximately matches cumulative pre-call reasoning tokens. The untouched parts of each demonstration appear twice as often in the truncated condition. We run the same RL procedure and evaluate both checkpoints with tools enabled. This design fixes which tasks and tool interactions are shown while changing the pre-call text and its repetition during SFT.

\subsection{Evaluation with and without Tool Access (Q4)}
\label{app:tool_access}
We evaluate each post-RL checkpoint (\high{} and LF25) in both inference conditions on the same 6,431 questions. With tools enabled, Zoom and Code are available and the model may also answer directly. The interaction allows up to eight tool calls and 8,192 response tokens. With tools disabled, we remove the tool schemas. Questions, decoding settings, and benchmark scoring are held fixed. For each checkpoint and benchmark we subtract tool-off final-answer accuracy from tool-on accuracy. The pooled value uses all 6,431 questions, so a question on which the model answers directly still counts in both conditions. Table~\ref{tab:tool_availability_gain} reports the differences in percentage points.

\subsection{Judging Tool-Use Decisions (Q5)}
\label{app:rebuttal_behavior_protocol}
We draw from the tool-on MathVista testmini (1,000 questions) and WeMath (1,740 questions) runs. For each benchmark, we sample separately from \high{} and LF25 at SFT initialization and after 32 RL updates. A trajectory is eligible once it attempts a supported tool call. We sample without conditioning on whether execution succeeded or the final answer was correct. Each checkpoint can therefore contribute different questions.

For first-call justification, GPT-6.1-Sol sees the question and the history up to the first call. It assesses whether the task and the evidence then available warranted using a tool. For joint tool and argument appropriateness, the judge sees the history up to each observed call, including earlier observations, the chosen tool, and its arguments. A call receives credit only if the tool suits the next step and its arguments specify the relevant operation on the intended inputs. For Zoom, the selected crop must include the relevant visual evidence. For Code, the supplied computation must use appropriate inputs for the quantity sought. For both decisions we withhold the current call's return, later turns, the ground-truth answer, and the checkpoint identity. For output use we show the successful Code return and the model's subsequent reasoning or answer. A positive judgment requires correct interpretation and use of that return. Failed executions provide no successful value to assess, and Zoom images are outside this last metric.

In Table~\ref{tab:rebuttal_behavior}, the denominator for each rate is, respectively, the number of sampled tool-using trajectories, the number of observed calls, or the number of successful Code calls. A trajectory with several calls contributes once to the first rate and multiple times to the second. These rates characterize observed decisions. They do not measure how often a model answered directly when a tool would have helped.

\section{Robustness across Seeds, Training Length, and Model Backbone}
\label{app:rebuttal_robustness}
Each comparison uses the full \high{} SFT data for one initialization and LF25 for the other. We report post-RL tool-on final-answer accuracy on V*Bench, ChartQA, MathVista, WeMath, and CharXiv. Overall accuracy pools questions across benchmarks. We compare initializations within each setting.

\subsection{Three Independent Training Seeds}
\label{app:seed_robustness}
For Qwen2.5-VL-7B-Instruct, we repeat each initialization under three independent training seeds, including the main run. Every run uses the 32-update GRPO configuration in Appendix~\ref{app:rl_config}, with 512 questions per update, five rollouts per question, and a reward based on final-answer correctness. We score each final checkpoint on the fixed tool-on suite, then compute the mean and sample standard deviation of the three accuracies for each initialization. LF25 leads in pooled accuracy in each run. Its mean is 69.58\% (SD 0.19 points), compared with 67.47\% (SD 1.08 points) for \high{} (Table~\ref{tab:rebuttal_seeds}).

\begin{table}[htbp]
\centering
\small
\setlength{\tabcolsep}{4pt}
\renewcommand{\arraystretch}{1.12}
\caption{Post-RL tool-on accuracy (\%), mean $\pm$ sample standard deviation across three seeds.}
\label{tab:rebuttal_seeds}
\begin{tabularx}{\linewidth}{@{}L{1.120000}C{0.980000}C{0.980000}C{0.980000}C{0.980000}C{0.980000}C{0.980000}@{}}
\toprule
\textbf{Initialization} & \textbf{V*Bench} & \textbf{ChartQA} & \textbf{MathVista} & \textbf{WeMath} & \textbf{CharXiv} & \textbf{Overall} \\
\midrule
\high{} & 83.42 \ensuremath{\pm} 1.51 & 86.15 \ensuremath{\pm} 0.75 & 66.10 \ensuremath{\pm} 1.05 & 58.22 \ensuremath{\pm} 1.13 & 35.20 \ensuremath{\pm} 3.36 & 67.47 \ensuremath{\pm} 1.08 \\
LF25 & \textbf{85.51 \ensuremath{\pm} 1.32} & \textbf{87.56 \ensuremath{\pm} 0.24} & \textbf{68.73 \ensuremath{\pm} 0.38} & \textbf{59.54 \ensuremath{\pm} 1.01} & \textbf{39.90 \ensuremath{\pm} 1.31} & \textbf{69.58 \ensuremath{\pm} 0.19} \\
\bottomrule
\end{tabularx}
\end{table}

\subsection{Extending RL from 32 to 64 Updates}
\label{app:longer_rl}
We extend Qwen2.5-VL-7B training for both initializations to 64 GRPO updates under the same reward, rollout budget, and optimizer settings used in the 32-update experiment. We compare the step-64 checkpoints on the five benchmarks. LF25 retains a 3.22-point pooled lead (Table~\ref{tab:rebuttal_64_accuracy}). Separately, we track the proportion of questions on which each checkpoint initiates a tool call at the start of training and at steps 8, 16, 24, 32, 48, and 64 (Table~\ref{tab:rebuttal_64_initiation}). By step 64 initiation has fallen to 0.12\% for \high{}. LF25 also drops early, then rises from 5.20\% at step 32 to 19.10\% at step 64.

\begin{table}[htbp]
\centering
\small
\setlength{\tabcolsep}{4pt}
\renewcommand{\arraystretch}{1.12}
\caption{Post-RL tool-on accuracy (\%) after 64 updates.}
\label{tab:rebuttal_64_accuracy}
\begin{tabularx}{\linewidth}{@{}L{1.750000}C{0.875000}C{0.875000}C{0.875000}C{0.875000}C{0.875000}C{0.875000}@{}}
\toprule
\textbf{Initialization} & \textbf{V*Bench} & \textbf{ChartQA} & \textbf{MathVista} & \textbf{WeMath} & \textbf{CharXiv} & \textbf{Overall} \\
\midrule
\high{} & 83.77 & 86.16 & 65.30 & 50.86 & \textbf{40.10} & 66.13 \\
LF25 & \textbf{84.29} & \textbf{88.12} & \textbf{68.90} & \textbf{58.05} & 39.70 & \textbf{69.35} \\
\bottomrule
\end{tabularx}
\end{table}

\begin{table}[htbp]
\centering
\small
\setlength{\tabcolsep}{4pt}
\renewcommand{\arraystretch}{1.12}
\caption{Checkpoint-level tool initiation rate (\%) in the extended RL run.}
\label{tab:rebuttal_64_initiation}
\begin{tabularx}{\linewidth}{@{}L{1.440000}C{0.937143}C{0.937143}C{0.937143}C{0.937143}C{0.937143}C{0.937143}C{0.937143}@{}}
\toprule
\textbf{Initialization} & \textbf{Start} & \textbf{Step 8} & \textbf{Step 16} & \textbf{Step 24} & \textbf{Step 32} & \textbf{Step 48} & \textbf{Step 64} \\
\midrule
\high{} & 59.80 & 12.34 & 6.29 & 4.92 & 2.11 & 0.20 & 0.12 \\
LF25 & 58.48 & 36.02 & 2.89 & 7.11 & 5.20 & 8.79 & 19.10 \\
\bottomrule
\end{tabularx}
\end{table}

\subsection{Changing the Backbone to Qwen3.5-4B}
\label{app:qwen35_backbone}
We train two Qwen3.5-4B models, one with the full \high{} SFT demonstrations and one with LF25, and evaluate both checkpoints after 32 RL updates. Both policies have access to Zoom and Code during evaluation. We compare their final-answer accuracies on the same five benchmarks, counting direct answers as well as tool-assisted answers. Table~\ref{tab:rebuttal_backbone} reports this comparison within Qwen3.5-4B. LF25 improves accuracy on all five benchmarks and raises the pooled score from 72.67\% to 74.35\%, a gain of 1.68 percentage points.

\begin{table}[H]
\centering
\small
\setlength{\tabcolsep}{4pt}
\renewcommand{\arraystretch}{1.12}
\caption{Qwen3.5-4B tool-on accuracy after 32 RL updates (\%).}
\label{tab:rebuttal_backbone}
\begin{tabularx}{\linewidth}{@{}L{1.750000}C{0.875000}C{0.875000}C{0.875000}C{0.875000}C{0.875000}C{0.875000}@{}}
\toprule
\textbf{Initialization} & \textbf{V*Bench} & \textbf{ChartQA} & \textbf{MathVista} & \textbf{WeMath} & \textbf{CharXiv} & \textbf{Overall} \\
\midrule
\high{} & 79.06 & 86.28 & 71.70 & 65.71 & 50.50 & 72.67 \\
LF25 & \textbf{81.68} & \textbf{86.84} & \textbf{73.70} & \textbf{67.05} & \textbf{55.10} & \textbf{74.35} \\
\bottomrule
\end{tabularx}
\end{table}

\section{SFT Data Composition and Tool-Specific Training Dynamics}
\label{app:data_and_dynamics}

\subsection{Data Characterization}
\label{app:rebuttal_data}
We report assistant-supervised trajectory lengths, full multimodal and text-only token allocation, and the source and call-depth compositions of the SFT data. Tool observations provide conditioning context and are masked from the SFT loss. The filtering score is average pre-call reasoning length per tool call rather than total trajectory length.
\begin{table}[htbp]
\centering
\small
\setlength{\tabcolsep}{4pt}
\renewcommand{\arraystretch}{1.12}
\caption{Distribution of assistant-supervised tokens per trajectory. P0/P100 are the minimum/maximum; P25, P50, and P75 are quantiles.}
\label{tab:rebuttal_lengths}
\begin{tabularx}{\linewidth}{@{}L{1.750000}C{0.875000}C{0.875000}C{0.875000}C{0.875000}C{0.875000}C{0.875000}@{}}
\toprule
\textbf{Dataset} & \textbf{Mean} & \textbf{P0} & \textbf{P25} & \textbf{P50} & \textbf{P75} & \textbf{P100} \\
\midrule
\low{} & 804.3 & 26 & 200 & 525 & 1,358.25 & 5,058 \\
\medium{} & 863.4 & 26 & 349.75 & 753.5 & 1,398 & 2,749 \\
\high{} & 930.8 & 26 & 496 & 921 & 1,319 & 3,867 \\
LF50 & 1,220.1 & 27 & 807 & 1,275 & 1,726.75 & 3,059 \\
LF25 & 1,353.3 & 28 & 929 & 1,354.5 & 1,945.25 & 3,059 \\
\bottomrule
\end{tabularx}
\end{table}

\begin{table}[htbp]
\centering
\small
\setlength{\tabcolsep}{4pt}
\renewcommand{\arraystretch}{1.12}
\caption{Native interaction token allocation (\%), including Zoom image tokens.}
\label{tab:rebuttal_native_tokens}
\begin{tabularx}{\linewidth}{@{}L{0.700000}C{1.100000}C{0.850000}C{1.000000}C{1.350000}@{}}
\toprule
\textbf{Dataset} & \textbf{Pre-call reasoning} & \textbf{Tool action} & \textbf{Tool observation} & \textbf{Final reasoning + answer} \\
\midrule
\low{} & 47.08\% & 9.60\% & 18.94\% & 24.39\% \\
\medium{} & 40.29\% & 10.54\% & 36.73\% & 12.44\% \\
\high{} & 26.44\% & 9.34\% & 59.34\% & 4.88\% \\
LF50 & 27.05\% & 6.93\% & 61.71\% & 4.31\% \\
LF25 & 31.90\% & 7.66\% & 54.52\% & 5.92\% \\
\bottomrule
\end{tabularx}
\end{table}

\begin{table}[htbp]
\centering
\small
\setlength{\tabcolsep}{4pt}
\renewcommand{\arraystretch}{1.12}
\caption{Text-only token allocation (\%), excluding image-token expansion.}
\label{tab:rebuttal_text_tokens}
\begin{tabularx}{\linewidth}{@{}L{0.700000}C{1.100000}C{0.850000}C{1.000000}C{1.350000}@{}}
\toprule
\textbf{Dataset} & \textbf{Pre-call reasoning} & \textbf{Tool action} & \textbf{Tool observation} & \textbf{Final reasoning + answer} \\
\midrule
\low{} & 56.57\% & 11.52\% & 3.04\% & 28.87\% \\
\medium{} & 60.35\% & 15.82\% & 5.50\% & 18.33\% \\
\high{} & 59.04\% & 21.08\% & 9.19\% & 10.68\% \\
LF50 & 65.52\% & 16.83\% & 7.42\% & 10.23\% \\
LF25 & 65.61\% & 15.77\% & 6.60\% & 12.02\% \\
\bottomrule
\end{tabularx}
\end{table}

We separately describe source-group composition and tool-type mixture. Across \high{}, LF50, and LF25, the reported tool-type proportions are 37.5\% Zoom, 37.5\% Code, and 25\% tool-free trajectories. Filtering nevertheless changes source composition and the call-depth distribution; average depth among tool-using trajectories decreases from 5.02 to 4.82.
\begin{table}[htbp]
\centering
\small
\setlength{\tabcolsep}{4pt}
\renewcommand{\arraystretch}{1.12}
\caption{Reported source-group distribution (\%).}
\label{tab:rebuttal_sources}
\begin{tabularx}{\linewidth}{@{}L{1.600000}C{0.800000}C{0.800000}C{0.800000}@{}}
\toprule
\textbf{Source group} & \textbf{\high{}} & \textbf{LF50} & \textbf{LF25} \\
\midrule
cof.zoom & 15.20\% & 3.47\% & 2.19\% \\
deepeyesv2.compute\_only & 15.70\% & 14.17\% & 16.88\% \\
mini\_o3.zoom & 34.80\% & 45.00\% & 39.38\% \\
thyme.compute\_only & 34.30\% & 37.36\% & 41.56\% \\
\bottomrule
\end{tabularx}
\end{table}

\begin{table}[htbp]
\centering
\small
\setlength{\tabcolsep}{4pt}
\renewcommand{\arraystretch}{1.12}
\caption{Tool-call depth distribution (\% of all trajectories). Mean depth excludes tool-free examples.}
\label{tab:rebuttal_depth}
\begin{tabularx}{\linewidth}{@{}L{1.600000}C{0.800000}C{0.800000}C{0.800000}@{}}
\toprule
\textbf{Number of calls} & \textbf{\high{}} & \textbf{LF50} & \textbf{LF25} \\
\midrule
0 (tool-free) & 25.00\% & 25.00\% & 25.00\% \\
4 & 35.95\% & 40.28\% & 45.62\% \\
5 & 18.60\% & 13.06\% & 8.44\% \\
6 & 10.70\% & 11.25\% & 11.56\% \\
7 & 6.05\% & 7.22\% & 8.12\% \\
8+ & 3.70\% & 3.19\% & 1.25\% \\
Mean depth (tool-using only) & 5.02 & 4.97 & 4.82 \\
\bottomrule
\end{tabularx}
\end{table}

\subsection{Tool-Specific Diagnostics}
\label{app:rebuttal_tools}
The longer pre-call supervision produced by filtering is present for both tools. From \high{} to LF25, the mean increases from 184 to 297 tokens for Zoom and from 134 to 216 for Code. The RL start/end summaries below further distinguish first and later calls: both tools exhibit longer first-call reasoning, while changes at later call positions depend on the tool and initialization. Tool initiation decreases for both Zoom and Code in these summaries, with a larger reduction for Zoom.
\begin{table}[htbp]
\centering
\small
\setlength{\tabcolsep}{4pt}
\renewcommand{\arraystretch}{1.12}
\caption{Mean pre-call reasoning tokens per call in the SFT datasets.}
\label{tab:rebuttal_sft_precall}
\begin{tabularx}{\linewidth}{@{}L{1.600000}C{0.800000}C{0.800000}C{0.800000}@{}}
\toprule
\textbf{Dataset} & \textbf{Overall} & \textbf{Zoom} & \textbf{Code} \\
\midrule
Original \high{} & 161 & 184 & 134 \\
LF50 & 231 & 262 & 189 \\
LF25 & 263 & 297 & 216 \\
\bottomrule
\end{tabularx}
\end{table}

\begin{table}[htbp]
\centering
\small
\setlength{\tabcolsep}{4pt}
\renewcommand{\arraystretch}{1.12}
\caption{Reported mean pre-call reasoning tokens at RL start $\rightarrow$ end, by tool and call position.}
\label{tab:rebuttal_rl_precall}
\begin{tabularx}{\linewidth}{@{}L{1.200000}C{0.650000}C{1.050000}C{1.050000}C{1.050000}@{}}
\toprule
\textbf{Initialization} & \textbf{Call position} & \textbf{Overall} & \textbf{Zoom} & \textbf{Code} \\
\midrule
\high{} & 1st & 266 \ensuremath{\rightarrow} 385 & 196 \ensuremath{\rightarrow} 348 & 284 \ensuremath{\rightarrow} 391 \\
\high{} & 2nd & 147 \ensuremath{\rightarrow} 162 & 202 \ensuremath{\rightarrow} 256 & 130 \ensuremath{\rightarrow} 142 \\
\high{} & 3rd & 156 \ensuremath{\rightarrow} 179 & 226 \ensuremath{\rightarrow} 322 & 126 \ensuremath{\rightarrow} 139 \\
LF25 & 1st & 312 \ensuremath{\rightarrow} 377 & 231 \ensuremath{\rightarrow} 354 & 339 \ensuremath{\rightarrow} 378 \\
LF25 & 2nd & 185 \ensuremath{\rightarrow} 202 & 252 \ensuremath{\rightarrow} 256 & 158 \ensuremath{\rightarrow} 197 \\
LF25 & 3rd & 195 \ensuremath{\rightarrow} 198 & 280 \ensuremath{\rightarrow} 268 & 147 \ensuremath{\rightarrow} 187 \\
\bottomrule
\end{tabularx}
\end{table}

\begin{table}[htbp]
\centering
\small
\setlength{\tabcolsep}{4pt}
\renewcommand{\arraystretch}{1.12}
\caption{Reported tool-specific initiation-rate summaries (\%) at RL start $\rightarrow$ end.}
\label{tab:rebuttal_tool_initiation}
\begin{tabularx}{\linewidth}{@{}L{1.600000}C{0.800000}C{0.800000}C{0.800000}@{}}
\toprule
\textbf{Initialization} & \textbf{Overall} & \textbf{Zoom} & \textbf{Code} \\
\midrule
\high{} & 62.27 \ensuremath{\rightarrow} 7.97 & 30.55 \ensuremath{\rightarrow} 1.68 & 32.66 \ensuremath{\rightarrow} 6.29 \\
LF25 & 57.15 \ensuremath{\rightarrow} 5.57 & 33.44 \ensuremath{\rightarrow} 0.22 & 26.02 \ensuremath{\rightarrow} 5.40 \\
\bottomrule
\end{tabularx}
\end{table}

\endgroup